\documentclass[preprint,12pt]{elsarticle}

\usepackage{graphicx}
\usepackage{subcaption}
\usepackage{hyperref}
\usepackage{makecell}
\usepackage[font=footnotesize,labelfont={bf,sc,small},figurename=Fig.,tablename=Tab.]{caption}

\hypersetup{
    colorlinks=true,
    linkcolor=blue.
    }

\usepackage{amssymb}

\usepackage{multirow}
\usepackage{color}
\usepackage[dvipsnames]{xcolor}
\usepackage{amsmath,amssymb,amsfonts}
\usepackage[algo2e,boxed,ruled,vlined]{algorithm2e} 

\definecolor{blue}{named}{black}

\newcommand{\eric}[1]{\textcolor{blue}{#1}}

\newcommand{\bb}[1]{\textcolor{blue}{#1}}

\begin{document}

\begin{frontmatter}



\title{Physics-Informed Echo State Networks for Modeling Controllable Dynamical Systems} 


\author[a]{Eric Mochiutti}
\author[a]{Eric Aislan Antonelo}
\author[a]{Eduardo Camponogara}

\affiliation[a]{organization={Department of Automation and Systems Engineering, Federal University of Santa Catarina},
            city={Florianópolis},
         postcode={88040-900}, 
            state={Santa Catarina},
          country={Brazil}}

\begin{abstract}
\eric{
Echo State Networks (ESNs) are recurrent neural networks usually employed for modeling nonlinear dynamic systems with relatively ease of training. 
By incorporating physical laws into the training of ESNs, Physics-Informed ESNs (PI-ESNs) were proposed initially to model chaotic dynamic systems without external inputs.
They require less data for training since Ordinary Differential Equations (ODEs) of the considered system help to regularize the ESN. 
In this work, the PI-ESN is extended with external inputs to model controllable nonlinear dynamic systems. 
Additionally, an existing self-adaptive balancing loss method is employed
to balance the contributions of the residual regression term and the physics-informed loss term in the total loss function.
The experiments with two nonlinear systems modeled by ODEs, the Van der Pol oscillator and the four-tank system, and with one differential-algebraic (DAE) system, an electric submersible pump,
revealed that the proposed PI-ESN outperforms the conventional ESN, especially in scenarios with limited data availability,
showing that PI-ESNs can regularize an ESN model with external inputs previously trained on just a few datapoints, reducing its overfitting and improving its generalization error (up to 92\% relative reduction in the test error).
Further experiments demonstrated that the proposed PI-ESN is robust to parametric uncertainties in the ODE equations and that model predictive control using PI-ESN outperforms the one using plain ESN, particularly when training data is scarce.
}

\end{abstract}




\end{frontmatter}







\section{Introduction}
\label{}



\eric{Physics-Informed Neural Networks (PINNs) \cite{Raissi2019, Karniadakis2021} combine physics laws into the training of neural networks, thereby reducing the need for labeled data points since the physics laws have a regularization effect on neural network training. In fact, PINNs can be trained using exclusively physics laws and initial or boundary conditions of 
Ordinary Differential Equations (ODEs) or Partial Differential Equations (PDEs) that describe real-world systems. These trained networks can then be used as proxy models for rapid simulation, showing computational speed, which are orders of magnitude better than numerical methods \cite{Raissi2019,Edwards2022,Eric2021}.
}

%
%


\eric{In the context of ODEs, PINNs map the continuous time $t$ input to an output  $\mathbf{y}(t)$ that are the states of the considered dynamical system as a function of time, representing the solution of the system's ODEs.
If a PINN is trained for a particular initial condition, and a specified range for the time $t$ input, then, after training, it will work only for that initial condition and specified time interval $[0,T]$.
%
Thus, if $t$ exceeds the values observed in the training data, PINNs tend to fail in their predictions.
An extension of PINNs is proposed in \cite{Eric2021}, called PINC (Physics-Informed Neural Nets for Control), to deal with variable longer time horizons that are not fixed in the training phase, by adding the initial state and control signal as additional input variables to the PINN. In addition, PINC makes PINN amenable to control applications.}

\eric{PINNs have been extended to other types of neural networks, such as 
Echo State Networks (ESNs) \citep{Jaeger2001a},
which are Recurrent Neural Networks with a simplified training scheme and one of the flavors of the
Reservoir Computing (RC) paradigm \cite{Verstraeten2006a}. 
ESNs have been used in several applications \cite{Antonelo2014,Zhou2022,ROBERTS2022,Jordanou2019} and achieve state-of-the-art 
performance in modeling chaotic dynamical systems \cite{SHAHI2022}. 
An ESN consists of a reservoir, a randomly generated RNN with fixed weights, and a linear adaptive readout output layer.
The reservoir projects the input into a high-dimensional nonlinear dynamic space. 
The instantaneous readings of the resulting reservoir state are further mapped to the target output using a linear readout output layer, the sole trainable part of the ESN.
This training is usually accomplished by one-shot linear regression techniques. 
Recent works have extended ESNs with residual connections in the temporal dimension \cite{Ceni2024residual},
and to deal with efficient processing of discrete-time dynamic temporal graphs \cite{Micheli2022discrete}.
In the current work, the ESN extension is called Physics-Informed ESNs (PI-ESNs) \cite{Doan2019}, which first trains the readout output layer with labeled data and then in a second stage with physics laws, leaving the reservoir untrained as usual.
PI-ESNs are originally used to model chaotic dynamical systems with no control inputs, so they can not readily be used for control applications. 
}
\eric{In this work, we propose a PI-ESN with control inputs, effectively showing that an ESN can be trained with physics laws and with randomly generated control input signals capable of leading the dynamical system to different operating points. 
Notably, we consider the small data regime, where only a few samples are labeled, i.e., have the target output, and the remaining unlabeled samples are used as collocation points in the physics law loss term
so that the output of the PI-ESN respects the physics laws of the considered system. 
Our proposed PI-ESN is the first ESN trained with differential equations of three systems that work with variable control inputs, making PI-ESN ready for control applications that employ Model Predictive Control (MPC) \cite{Camacho:2007}. 
Notice also that while PINNs explicitly accept a continuous time input, PI-ESNs do not require it since they treat time implicitly in a discrete way via the recurrent connections that form an internal memory in the reservoir. Compared to PINC \cite{Eric2021}, our PI-ESN loops at the reservoir level, while PINC feeds the predicted output of the states back as the initial condition of the next time interval. Thus, we consider our PI-ESN inherently more ready for MPC applications in discrete time.
}

\eric{
The contributions of this work are as follows: 
\begin{enumerate}
    \item 
  a PI-ESN architecture with external inputs is proposed to allow the modeling of controllable dynamical systems, differently from \cite{Doan2019}, which only tackles ESNs for autonomous systems without external inputs;
\item 
a self-adaptive balancing loss for PI-ESN is presented, balancing each term (data loss and physics-based loss) in the loss function dynamically, replacing a slow, manual trial and error procedure; 
the resulting scheme is called PI-ESN-a, inspired by \cite{Xiang2022} for PINNs, improving the training and the final performance of the network. For instance, PI-ESN-a improves by 30.9\% on an error metric compared to plain PI-ESN on the Van der Pol Oscillator.
\item an extensive set of experiments is presented showing the performance gain and predictive power of PI-ESN-a over the pure ESN in low-data regimes for three representative dynamical systems, one of them being an electric submersible pump, which has three states, two control inputs and seven algebraic variables. 
This shows that physics laws can regularize ESNs with external inputs as well. 
Note that more recent ESN architectures can also be incorporated into the PI-ESN-a framework, provided the proposed physics-based training methodology remains as outlined in this work. Thus, PI-ESN-a does not compete with newer ESN architectures but instead leverages unlabeled data and existing physical laws to enhance the performance of any specific ESN.
\item the proposed physics-informed ESN, PI-ESN-a, is used for the first time in the MPC of a dynamical system plant, the four-tank system which has multiple inputs and multiple outputs (MIMO). This is in contrast to PINC \cite{Eric2021}, which needs to feed back the predicted states as initial conditions in a self-loop scheme, and to plain ESN \cite{Jordanou2021} which needs more data points to train sufficiently well the network.
\end{enumerate}
}


\eric{
This work is structured as follows. Section ~\ref{sec:related} presents the related works. 
Section \ref{sec:methods} describes the proposed architecture, including the ESN, PI-ESN and self-balancing terms of the loss function.
Section \ref{sec:experiments} shows the application of method for the Van der Pol oscillator, for the four-tank system, and for the electric submersible pump model. It also provides an evaluation in scenarios characterized by limited availability of training data. 
Section \ref{sec:conclusion} concludes this work.
}

\section{Related Works}
\label{sec:related}
The work in \cite{Pathak_2018} 
deals with hybrid forecasting of chaotic processes, where
a hybrid scheme employs an approximate knowledge-based model jointly with an ESN (or Reservoir Computing network). Their hybrid scheme yields better state prediction performance for two applications, namely in forecasting the Lorenz system and the Kuramoto-Sivashinsky equations, both chaotic systems.
Basically, their method extends the input fed to the reservoir network to include the prediction given by the approximate knowledge-based model (whose physics equations are simulated with traditional methods). Thus, it makes it easier for the ESN \eric{because it is required} only to learn to correct the \textit{mistakes} of the approximate knowledge-based model. It is worth noting that \cite{Pathak_2018} does not include physics laws in the training of ESNs as PI-ESNs do, but it does require \eric{numerical} simulation of the model (unlike PI-ESNs, which requires only a few initial collected data). 
\eric{In addition, their application of ESN is limited to autonomous systems without external inputs, unlike our work.}

\eric{The PI-ESN approach was used for accurate prediction of extreme events and abrupt transitions in self-sustaining turbulence processes in \cite{doan20192}, where the physics modeling part captures conservation laws. 
Their application also considered only autonomous systems without external inputs.
 }
%

The PI-ESN framework is expanded to reconstruct the evolution of unmeasured states of chaotic systems in \cite{Doan2020}. By training the PI-ESN using data devoid of unmeasured state information and the physics equations of a chaotic system, the network can accurately reconstruct the unmeasured state. 
\eric{Their experiments focus on non-controllable dynamical systems without external input, unlike our work.}

The method proposed by \cite{Racca2021} represents an improvement over the previous approach by utilizing automatic differentiation for physics error computation, instead of relying on an explicit Euler integration scheme. Through the application of automatic differentiation, the author's method achieves a more precise estimation of the inherent physical error within the chaotic system. This heightened accuracy can be attributed to the meticulous treatment of gradients and derivatives, enabling a finer adjustment of model parameters during the training process.

\eric{In \cite{oh2023pure}, 
a pure physics-informed ESN is introduced,
where training consists of regression in two stages with the differential equation itself. Their method does not rely on labeled training data and is developed only for dynamic systems without external inputs.}

\eric{In contrast to the previous approaches, our proposed method's emphasis is on controllable systems driven by external input signals,
diverging from the approaches outlined in the other works, which present results only for autonomous dynamic systems without external inputs. 
Thus, our method can be used in control applications, e.g., MPC applications, where an ESN regularized by physics laws serves as a predictive model in an optimization loop to control a plant.
}

\section{Methods}
\label{sec:methods}

\subsection{Echo State Networks}
\label{subsec:esn}

\subsubsection{Model}

\begin{figure}[h!]
    \centering
    \includegraphics[width = 0.65\textwidth]{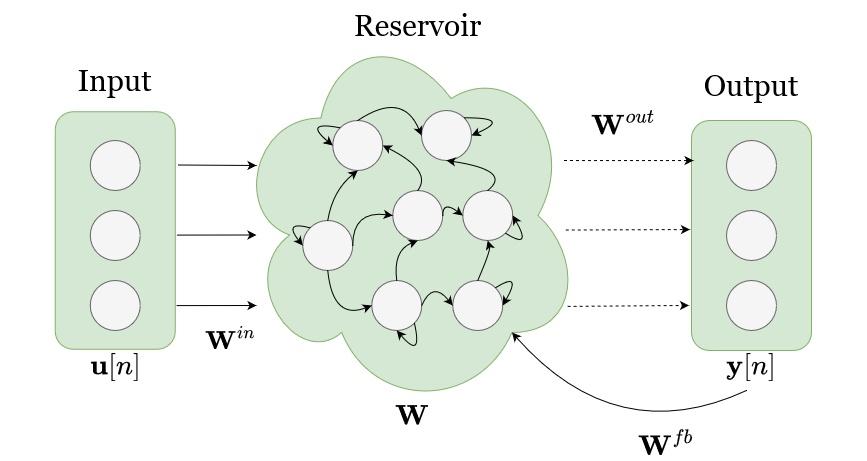}
    \caption{Echo State Network (ESN) architecture. 
    The reservoir is a recurrent neural network with inner weights and incoming weight connections (solid lines) that are all randomly generated and fixed, 
    projecting the input to a high-dimensional dynamic nonlinear space.
    A linear readout output layer linearly projects the reservoir states to the desired output (dotted lines).
    The output $ \mathbf{ y}$ can be fed back to the reservoir via $\mathbf{W}^{fb}$.
    }
    \label{fig:ESN_freerun}
\end{figure}

\eric{The standard ESN model with output feedback is shown in \autoref{fig:ESN_freerun}.} Given an input signal $\mathbf{u}[n] \in \mathbb{R}^{N_u}$ and the corresponding output signal $\mathbf{y}[n] \in \mathbb{R}^{N_y}$ for $n = 1,\hdots,N$ time steps,
the state update equation for the reservoir states
$\mathbf{x}[n] \in \mathbb{R}^{N_x}$ are as follows:
\begin{align}
\mathbf{x}[n + 1] &= (1- \alpha) \mathbf{x}[n] + \alpha f({\mathbf{W}^{in}}\mathbf{u}[n + 1] + \mathbf{W}\mathbf{x}[n] + {\mathbf{W}^{fb}}\mathbf{ y}[n]),
\label{eq:stateupESN}
\end{align}
where: $f$ is the activation function, usually $\tanh$;  $\alpha \in (0,1]$ is the leak rate \cite{Jaeger2001a};
$\mathbf{W}^{in} \in {\mathbb{R}^{N_x \times N_u}}$ represent the connections from input to the reservoir, $\mathbf{W} \in {\mathbb{R}^{N_x \times N_x}}$ are the recurrent connections in the reservoir, and
$\mathbf{W}^{fb} \in {\mathbb{R}^{N_x \times N_y}}$ represent the feedback connections from the readout output to the reservoir. All connections going to the reservoir ($\mathbf{W}^{in}$, $\mathbf{W}$, and $\mathbf{W}^{fb}$) are randomly initialized and fixed.

The readout output $\mathbf{y}[n+1]$ is given by:
\begin{align}
    \mathbf{y}[n + 1] &= \mathbf{W}^{out}\mathbf{x}[n+1],
\label{eq:outputESN}
\end{align}
where: $\mathbf{W}^{out} \in {\mathbb{R}^{N_y \times N_x}}$ are the adaptive weights of the readout output layer. 

When the value of $\alpha$ is closer to zero, it will make the reservoir dynamically slower, increasing ``memory'' over previous states \cite{Jaeger2001a}.
The non-trainable weights are set as follows. The input weights in $\mathbf{W}^{in}$ are $0$, $\delta_{in}$, and $-\delta_{in}$ with probabilities of 0.5, 0.25, and 0.25, respectively, where $\delta_{in}$ serves as an input scaling factor, while the feedback weights $\mathbf{W}^{fb}$, are chosen from a uniform distribution in the interval [$\delta_{fb}$, $-\delta_{fb}$]. Here, $\delta_{fb}$ represents the feedback scaling hyperparameter.
The values in $\mathbf{W}$ are randomly chosen from a uniform distribution in the interval $[-1, 1]$.
Afterward, the spectral radius $\rho(\mathbf{W}) = \max \{ |\lambda|$ : $\lambda$ is an eigenvalue of $\mathbf{W}\}$ is selected to ensure that $\rho(\mathbf{W})<1$. The spectral radius is usually selected as close as possible to 1, where the reservoir operates at the edge of stability. This is done to generate more diverse signals that can contribute to the identification of the dynamics of a system. 
Here, $\rho ^*$ is a hyperparameter that scales $\mathbf{W}$ so that its spectral radius is equal to the value of $\rho ^*$ \cite{Yildiz2012}:
\begin{equation}
{\mathbf{W}} = {\mathbf{W}^*}\frac{{{\rho ^*}}}{{\rho (\mathbf{W}^*)}},
\label{eq:spectralradius}
\end{equation}
where $\mathbf{W}^*$ represents random values before applying the desired spectral radius value.

\subsubsection{Training}

\eric{The output layer is trained so that the mean squared error $J_{data}$ is minimized \cite{Lukoševičius2012}:}
\begin{equation}
{J_{data}} = \frac{1}{{{N_y}}}\sum\limits_{i = 1}^{{N_y}} {\frac{1}{{{N_t}}}} \sum\limits_{n = 1}^{{N_t}} {{{\Bigl [ {{{\hat y}_i}[n] - {y_i}[n]} \Bigr ]}^2}},
\label{eq:Jdata}
\end{equation}
where: $N_t$ represents the number of training data samples;
and $N_y$ represents the output dimension. Ridge regression is usually employed to find ${\mathbf{W}^{out}}$:
\begin{equation}
     {\mathbf{W}^{out}} = \mathbf{\hat Y}{\mathbf{X}^T}{(\mathbf{X}{\mathbf{X}^T} + \gamma \mathbf{I})^{ - 1}},
    \label{eq:ridgeregression}
\end{equation}
where: $\mathbf{X}$ and $\mathbf{\hat Y}$ represent the column concatenation of the $N_t$ instants of the ESN states $\mathbf{x}[n]$ and corresponding desired output $\mathbf{\hat y}[n]$, respectively, defined in 
\autoref{eq:Xreservoir} and \autoref{eq:Ytarget}; 
and $\gamma$ is the Tikhonov regularization factor.
\begin{equation}
 \mathbf{X} \in {{\rm I\!R}^{{N_x} \times {N_t}}} = \left[ {\begin{array}{*{20}{c}}
{{x_1}[1]}&{\hdots}&{{x_1}[{N_t}]}\\
{\begin{array}{*{20}{c}}
\vdots
\end{array}}&{\begin{array}{*{20}{c}}
\ddots
\end{array}}&{\begin{array}{*{20}{c}}
\vdots
\end{array}}\\
{{x_{{N_x}}}[1]}&{\hdots}&{{x_{{N_x}}}[{N_t}]}
\end{array}} \right]
\label{eq:Xreservoir}
\end{equation}

\begin{equation}
\mathbf{\hat Y} \in {{\rm I\!R}^{{N_y} \times {N_t}}}  = \left[ {\begin{array}{*{20}{c}}
{{\hat y_1}[1]}&{\hdots}&{{\hat y_1}[{N_t}]}\\
{\begin{array}{*{20}{c}}
\vdots
\end{array}}&{\begin{array}{*{20}{c}}
\ddots
\end{array}}&{\begin{array}{*{20}{c}}
\vdots
\end{array}}\\
{{ \hat y_{{N_y}}}[1]}&{\hdots}&{{\hat y_{{N_y}}}[{N_t}]}
\end{array}} \right]
\label{eq:Ytarget}
\end{equation}

When generating $\mathbf{x}[n]$ to build the designed matrix $\mathbf{X}$ in \autoref{eq:Xreservoir}, the output is teacher-forced using the desired output $\mathbf{\hat{y}}[n]$,
i.e., 
instead of \autoref{eq:stateupESN}, during training, we use:
\begin{align}
\mathbf{x}[n + 1] &= (1- \alpha) \mathbf{x}[n] + \alpha f({\mathbf{W}^{in}}\mathbf{u}[n + 1] + \mathbf{W}\mathbf{x}[n] + {\mathbf{W}^{fb}}\mathbf{\hat{y}}[n])
\label{eq:stateupESN_teacherforce}
\end{align} 
This desired output $\mathbf{\hat{y}}[n]$ comes from the industrial plant or from a phenomenological model as collected data, for instance. 
%
After training, \autoref{eq:stateupESN_teacherforce} is used for a few initial timesteps to warm up the reservoir, and then \autoref{eq:stateupESN} is utilized normally, where the actual output prediction $\mathbf{y}[n]$ is fed back to the reservoir.


\subsubsection{\eric{Hyperparameter tuning}}
\label{}

To optimize the ESN, the data set $(1,\hdots, N_t)$ is divided into a training set ($1, \hdots, N_{te}$) and a validation set ($1 + N_{te}, \hdots, N_{t}$). Hence, $N_{te} + N_{ve} = N_t$, where $N_{te}$ represents the number of training data samples and $N_{ve}$ stands for the number of validation data samples used in the hyperparameter search. The optimization process employs a grid search to determine the best values for $\delta_{in}$, $\delta_{fb}$, and $\gamma$. Once the best values of hyperparameters are identified, the ESN is retrained using all available data ($1,\hdots, N_t$).

\subsection{Physics-Informed Neural Networks}
The work by \cite{Raissi2019} introduced the concept of Physics-Informed Neural Networks (PINNs), which involves training deep neural networks in a supervised manner to adhere to physical laws described by PDEs or ODEs. 
\eric{In this work, we consider nonlinear ODEs of the following general form:
}
\begin{equation}
   \partial_t \mathbf{y} - \mathcal{N}[\mathbf{y}]=0,  \quad t \in [0,T] \label{eq:general_ode}
\end{equation}
where $\mathcal{N}[\cdot]$ is a nonlinear differential operator and $\mathbf{y}$ represents the state of the dynamic system (the latent ODE solution).
\eric{We define $\mathcal{F}(\mathbf{y})$ to be equivalent to the left-hand side of Eq. \ref{eq:general_ode}:
}
\begin{equation}
\mathcal{F}(\mathbf{y}) \equiv \partial_t \mathbf{y} - \mathcal{N}[\mathbf{y}], 
\label{eq:fisicageral}
\end{equation}
\eric{Here, $\mathbf{y}$ also represents the output of a multilayer neural network (hence the notation $\mathbf{y}$ instead of $\mathbf{x}$) which has the continuous time $t$ as input.
%
This formulation implies that a neural network must learn to compute the solution of a given ODE.
Thus, when $\mathcal{F}(\mathbf{y}) = 0$, the neural network output respects the laws described by the ODEs perfectly for the considered time interval. }



Assuming an autonomous system for this formulation, a given neural network $\mathbf{y}(t)$ is trained using optimizers such as ADAM \citep{Kingma2014} or L-BFGS \citep{Andrew2007} to minimize
a mean squared error (MSE) cost function:
\begin{equation}
\textrm{MSE} = \textrm{MSE}y + \textrm{MSE}_{\mathcal{F}},
\label{eq:erro}
\end{equation}
Here, the first loss term $\textrm{MSE}y$ corresponds to the typical loss function for regression \citep{Bishop2006}, relying on training data to establish initial conditions for the ODE solutions. The second loss term $\textrm{MSE}_{\mathcal{F}}$ penalizes any discrepancies in the behavior of $\mathbf{y}(t)$, as gauged by $\mathcal{F}(\mathbf{y})$ in \autoref{eq:fisicageral}. This term enforces the physical nature of the solution by considering $\mathcal{F}(\mathbf{y})$ at a finite selection of randomly sampled collocation points.

\subsection{Physics-Informed Echo State Network with External Input (PI-ESN-i)}

\subsubsection{Architecture}

\begin{figure}[h!]
    \centering
    \includegraphics[width = 0.8\textwidth]{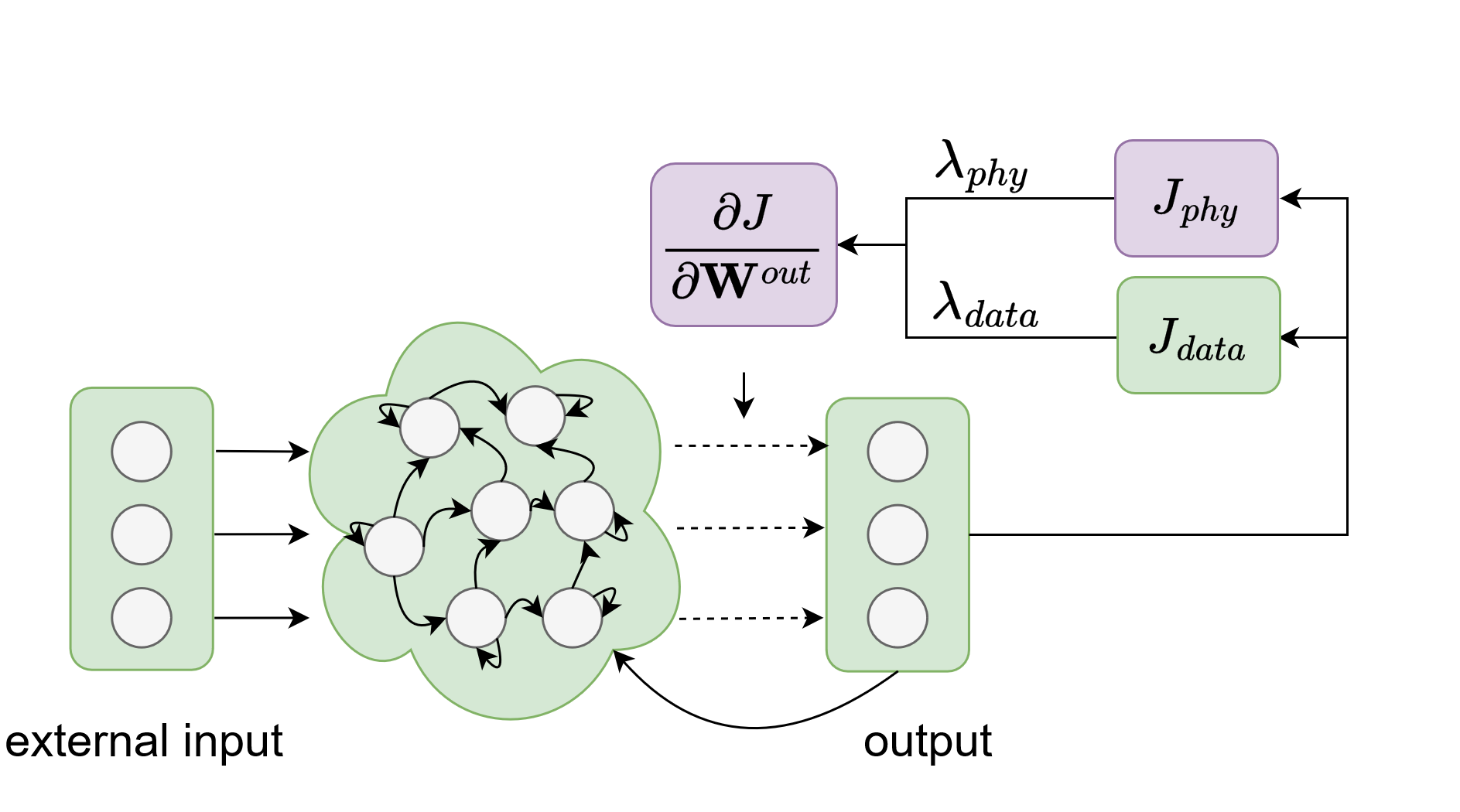}
    \caption{Physics-Informed Echo State Network with external Inputs \eric{(PI-ESN-i)}.
    While PINNs accept explicitly continuous time $t$ as input, PI-ESN-i treats time in a discrete way and implicitly by the recurrent reservoir updates.
    The ESN output is used to calculate the physics-informed loss function $J_{phy}$ using collocation points and the data loss function $J_{data}$ using data points. 
    These loss functions are scaled by $\lambda_{phy}$ and $\lambda_{data}$, respectively, to form the total loss $J$.
    The derivative of the loss ${\partial J}/{\partial \mathbf{W}^{out}}$ is calculated by automatic differentiation and used to update the values of the output layer $\mathbf{W}^{out}$.
    }
    \label{fig:PI_ESN}
\end{figure}

A traditional PINN needs \eric{a continuous time $t$ as input and does not inherently  work with external control inputs.
} 
The time input does not exist in the PI-ESN since an ESN is a discrete-time recurrent network that implicitly incorporates time through the network state update equation (\autoref{eq:stateupESN}), where the next state depends on the previous state.
%
In this work, we extend the PI-ESN to accept external inputs \eric{as shown in \autoref{fig:PI_ESN}}, such as plant control inputs in addition to the output feedback itself.
As a recurrent neural network with external inputs, the PI-ESN-i architecture can simulate for an arbitrary period of time and is ready to be used in control applications, if desired, unlike conventional PINNs.

\subsubsection{Training}
%
%

In a PI-ESN-i, $\mathbf{W}^{out}$ will be adapted following physics laws described by ODEs or DAEs (Differential-Algebraic Equations). The initial estimate for $\mathbf{W}^{out}$ is computed by ridge regression as in \autoref{eq:ridgeregression} on the available training data. 
We assume that this dataset is limited in size such that additional physics-informed training will be beneficial to further improve the ESN prediction accuracy and generalization.

This training data consist of $N_t$ data points, $\{(\mathbf{u}[n],\mathbf{\hat{y}}[n]), n=1,\cdots,N_t \}$, in the time interval $[0,T]$.
The loss function for the training data, $J_{data}$, is given by equation \autoref{eq:Jdata}.
Notice that this loss function is minimized during pretraining of $\mathbf{W}^{out}$ through Ridge Regression in one-shot learning and also iteratively in conjunction with the physics-informed training, as we will see below.

\begin{figure}[h!]
    \centering
    \includegraphics[width = 0.9\textwidth]{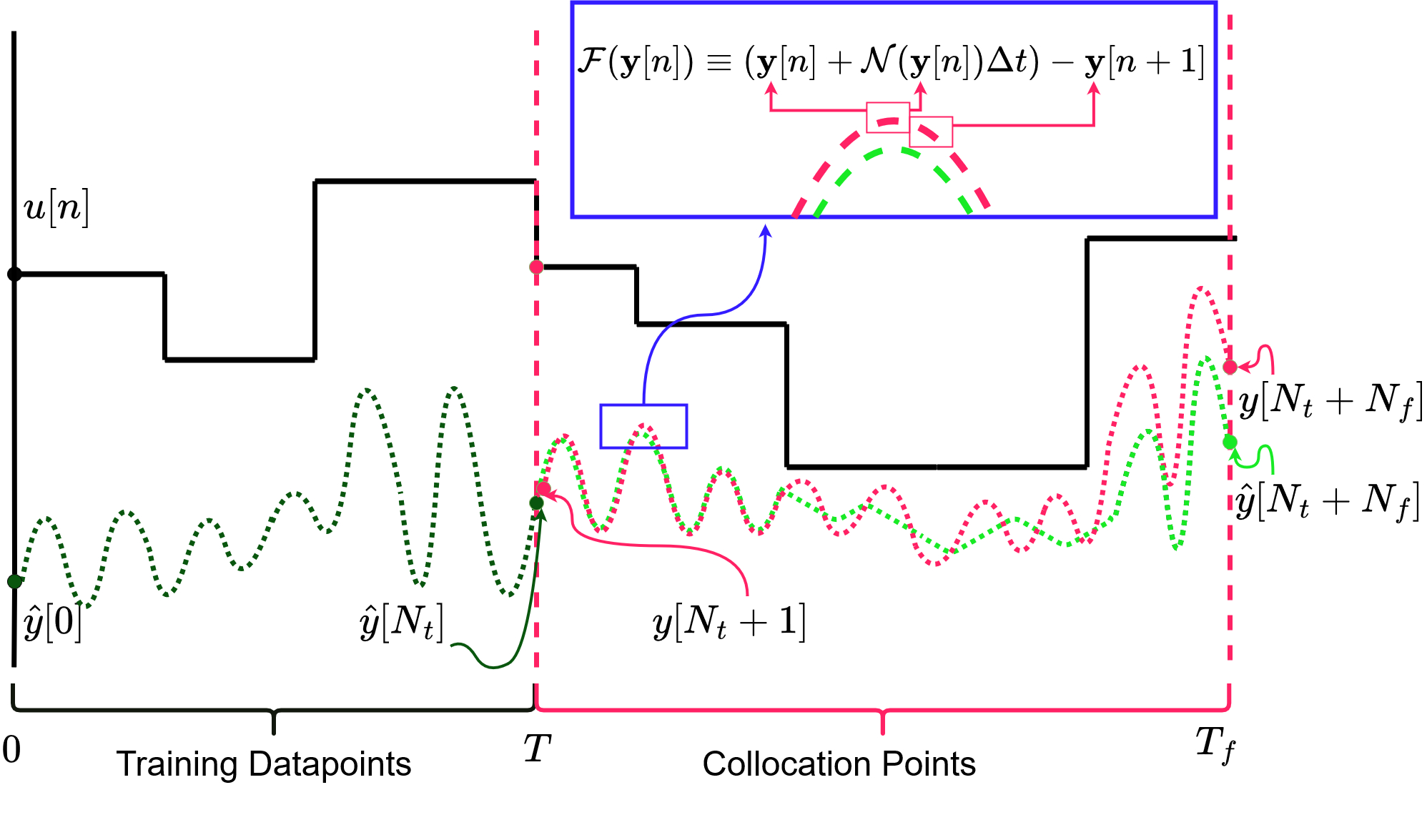}
    \caption{
    %
    \eric{
    Representation of the PI-ESN-i samples used to calculate the total loss function. 
    In solid black, we have the input data $\mathbf{u}[n]$, which can be multidimensional. 
    The desired output is given by the dashed green line, while the ESN's output prediction is given by the dashed pink line.
The first section (training datapoints, $t \in [0, T]$) is used in pretraining the ESN's output layer and also in
the data loss term included in the total loss function. 
The second section (collocation points, $t \in [T, T_f]$) is used in the physics-informed loss term included in the total loss function, and training only uses the output prediction $\mathbf{y}[n]$ to compute the loss \autoref{esn:eq:custofisico}, as indicated in the blue box. 
The target $\mathbf{\hat{y}}[n]$ in light green is assumed to be unknown for $t > T$, but it is shown to illustrate that $\mathbf{y}[n]$ and $\mathbf{\hat{y}}[n]$ mismatch prior to training, and a performance measurement can be computed after training if it is available.}
    %
}
    \label{fig:Data_collocation}
\end{figure}

Enforcing the physics of the underlying system on the ESN means that its output $\mathbf{y}[n]$ should satisfy the ordinary differential equations (ODE) of the considered system or plant for the entire time horizon of interest, denoted as $t \in (T, T_f]$. 
In this time interval, which is after $t=T$ or $N_t$ labeled data points have been collected, the desired output $\mathbf{\hat y}[n]$ is unavailable. However, physics laws can still be evaluated and enforced on the ESN's outputs during iterative training for 
$N_f$ collocation points drawn from time interval $(T, T_f]$, i.e., 
$\{\mathbf{u}[n], n=N_t+1,\cdots,N_t+N_f \}$, are generated temporally right after the training data, where $\Delta t $ is the discrete time sampling interval for the samples in $(T, T_f]$. These collocation points will be used to enforce the physics on the output $\mathbf{y}[n]$ of the ESN. \autoref{fig:Data_collocation} shows an illustration of this process.

Notice that \autoref{eq:stateupESN_teacherforce} and \autoref{eq:Xreservoir}, \eric{which  employ teacher-forcing of the desired output}, are used to generate $\mathbf{X}_t$, i.e., the states during the interval $[0, T]$  (training data). On the other hand,
\autoref{eq:stateupESN}, \eric{which feeds back the output's prediction}, is applied to generate
 $\mathbf{X}_f$, which corresponds to the states in $(T, T_f]$ (collocation points):
 
 \begin{equation}
 \mathbf{X}_f \in {{\rm I\!R}^{{N_x} \times {N_f}}} = \left[ {\begin{array}{*{20}{c}}
{{x_1}[N_t+1]}&{\hdots}&{{x_1}[{N_t + N_f}]}\\
{\begin{array}{*{20}{c}}
\vdots
\end{array}}&{\begin{array}{*{20}{c}}
\ddots
\end{array}}&{\begin{array}{*{20}{c}}
\vdots
\end{array}}\\
{{x_{{N_x}}}[N_t+1]}&{\hdots}&{{x_{{N_x}}}[{N_t + N_f}]}
\end{array}} \right]
\label{eq:Xf}
\end{equation}
This is because during $[0, T]$, the desired output $\mathbf{\hat{y}}[n]$ is known and, thus, it is teacher-forced. For the other interval with the collocation points, the ESN's output prediction $\mathbf{y}[n]$ is fed back instead, meaning the ESN is in free-run mode. This can cause some instabilities during training if the output is fed back at each weight update.

While $\mathbf{X}_t$ is used both in pre-training and physics-informed training of $\mathbf{W}^{out}$ and is kept fixed during the whole training process,
 $\mathbf{X}_f$ is exclusively used in physics-informed training and can change as training evolves, \eric{since it is computed using the output's prediction}. How often we update $\mathbf{X}_f$ will influence the training convergence.


\eric{The network's output $\mathbf{y}[n]$ aims to approximate the states of a dynamic system described by ODEs and is used to calculate the physical error $\mathcal F$, as described in \autoref{eq:fisicageral}.}
Typically, for conventional PINNs, the derivative $\partial_t \mathbf{y}$ is obtained through automatic differentiation with respect to its time input. In the case of PI-ESN, as it lacks an explicit time input and operates in discrete time,  \autoref{eq:fisicageral} needs to be discretized using numerical methods like explicit Euler or Runge-Kutta \cite{Doan2019}, for example. Using the first method and considering that $\mathcal{F}(\mathbf{y}) = 0$, we have the following:
\begin{equation}
\mathcal F(\mathbf{y}) \equiv \mathbf{y}[n+1] - (\mathbf{y}[n] + \mathcal N(\mathbf{y}) \Delta t)
\label{eq:euler}
\end{equation}

This function $\mathcal{F}$ is applied at each collocation point. In other words, using the state matrix $\mathbf{X}_f$ related to the second interval $t \in (T, T_f]$, we calculate the physics-related loss $J_\mathrm{physics}(\mathbf{W}^{out})$:

\begin{equation}
J_\mathrm{physics}(\mathbf{W}^{out})
= \frac{1}{{{N_y}}}\sum\limits_{i = 1}^{Ny} {\frac{1}{{{N_f}}}\sum\limits_{n = N_t+1}^{{N_f}} {{{\left| {\mathcal{F}({y_i}[n])} \right|}^2}} } 
\label{esn:eq:custofisico}
\end{equation}

The total loss function to be minimized takes into account both the data loss and the residual function from the physics laws:
\begin{equation}
    J(\mathbf{W}^{out}) = \lambda_{data} \cdot J_\mathrm{data}(\mathbf{W}^{out}) + {\lambda _{phy}} \cdot J_\mathrm{physics}(\mathbf{W}^{out}),
    \label{esn:eq:custototal}
\end{equation}
\noindent
where ${\lambda _{data}}$ and ${\lambda _{phy}}$ are hyperparameters used to balance the importance of the two terms in the loss functions during optimization. 
Typically, the hyperparameter ${\lambda_{data}}$ is set to a default value of 1, while only ${\lambda_{phy}}$ is adjusted to define the relative importance between the loss terms. 
The total loss function $J(\mathbf{W^{out}})$ in \autoref{esn:eq:custototal} is minimized by iterative update of $\mathbf{W^{out}}$, employing ADAM or L-BFGS optimizers available in frameworks like TensorFlow or PyTorch.
It is important to remember that automatic differentiation does not pass through the recurrence of the reservoir nor through the output feedback; that is, there is no backpropagation of errors through time. Thus, the gradient of $J(\mathbf{W^{out}})$ is computed by considering that past terms are not dependent on $\mathbf W^{out}$.
%
Since $\mathbf{X}_f$ depends on $\mathbf{W}^{out}$
because there is feedback from the previous output $\mathbf{y}[n]$ in the states $\mathbf{x}[n+1]$, the values of all states would change at each weight update (of $\mathbf{W}^{out}$). 
To prevent this from happening, the states are only updated with the new feedback every $K$ iterations of the optimizer in an attempt to stabilize the training process.
 This is equivalent to using two versions of $\mathbf{W}^{out}$, one that is constantly updated by training and another used to calculate the ESN output 
 which is updated only every $K$ iterations to the value of the first\footnote{This setup with two weights matrices is usually done on Deep Reinforcement Learning as well.}. 
 Algorithm \ref{algor:adaptive_PI_ESN} presents a pseudocode \eric{for training the PI-ESN with external inputs}.

\subsection{Adaptive Balanced Loss for PI-ESN-i}

It is not straightforward to balance the physics-informed loss and data loss terms in \autoref{esn:eq:custototal}. The scalings $\lambda_{physics}$ and $\lambda_{data}$ that balance the loss are hyperparameters whose values depend on the particular application (the system being modeled). Besides, once chosen, these scaling values stay fixed throughout training, which may not be ideal.
In \cite{Xiang2022}, PINNs are augmented with adaptive $\lambda_{physics}$ and $\lambda_{data}$  hyperparameters, which are updated together with the PINN's weights in the optimization procedure. Thus, the balance between the data and the physic losses evolves dynamically with the training process while
relying solely on the training data that is available initially.

\begin{figure}[h!]
    \centering
    \includegraphics[width = 0.85\textwidth]{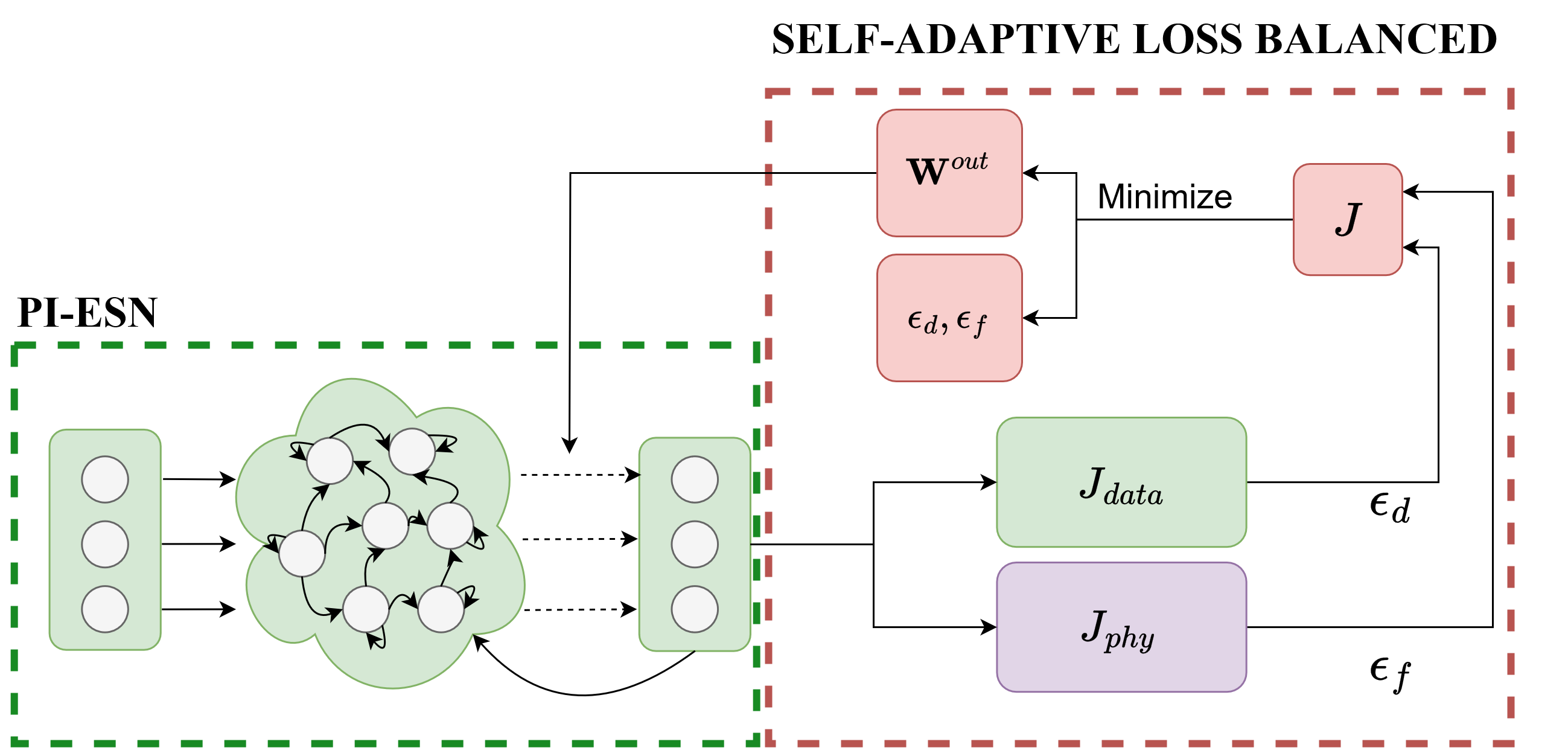}
    \caption{Physics-Informed Echo State Network with \eric{external Inputs and} Self-Adaptive Balancing Loss (PI-ESN-a). 
    The loss function of PI-ESN-a has adaptive scaling parameters
    $\epsilon_d$ and $\epsilon_f$ that dynamically balance the contributions of 
    $J_{data}$ and $J_{phy}$ into the total loss function (\autoref{esn:eq:gaussianlikelihood_2}), respectively.
    %
    An optimizer is employed to minimize this function and
    \eric{update $\mathbf{W}^{out}$ and the adaptive parameters $\epsilon_d$ and $\epsilon_f$. }
    This optimization procedure is outlined in \autoref{algor:adaptive_PI_ESN}.}
    \label{fig:selfadaptive_PI_ESN}
\end{figure}

In this work, our PI-ESN-i is augmented similarly to the conventional PINN in \cite{Xiang2022} with a self-adaptive balancing loss that dynamically balances both terms in the loss (\autoref{fig:selfadaptive_PI_ESN}), noted as PI-ESN-a.
Our goal is to maximize the Gaussian log-likelihood of a probabilistic model based on the ESN's output. Thus, we consider that the ESN's output $y(\mathbf{u}; \mathbf{W}^{out})$\footnote{The output $y$ is a function of the input vector $\mathbf u$ and implicitly of the internal reservoir state $\mathbf{x}$, and parametrized by the adaptive $\mathbf{W}^{out}$.
\eric{For the sake of simplicity, the notation for $\mathbf u$ and $\mathbf x$ includes all timesteps and not just a single one.}
} 
parametrizes a Gaussian distribution
\begin{equation}
p( \Tilde{y} \mid y(\mathbf{u}; \mathbf{W}^{out})) =
\mathcal{N} (y(\mathbf{u}; \mathbf{W}^{out}), \epsilon_d ^2)
\label{esn:eq:gaussianlikelihood}
\end{equation}
where: 
\eric{$\Tilde{y}$ is the observed output;} and
the mean of the Gaussian is given by the ESN's output, and its standard deviation $\epsilon_d$ defines the uncertainty of the model for the training data. This uncertainty $\epsilon_d$ is usually fixed in weight decay or regularization of neural networks, but here, it will be tuned by maximum likelihood inference. This is equivalent to minimizing the negative log-likelihood of the model:
    \begin{equation}
\begin{aligned}
-\log p( \Tilde{y} \mid y(\mathbf{u}; \mathbf{W}^{out}) )
&
\propto 
\frac{1}{{2{\epsilon^2 _d}}}  \left\| \Tilde y - y(\mathbf{u}; \mathbf{W}^{out}) \right\|^2 
    + 
\log {\epsilon_d}
\\
& = 
\frac{1}{{2{\epsilon^2 _d}}} J_{data} (\mathbf{W}^{out})
+
  \log {\epsilon_d}
  \end{aligned}\label{esn:eq:gaussianlikelihood}
  \end{equation}

Now, we can add another output $h$ to the Gaussian probability model that represents the physics law applied to the ESN's output $y(\mathbf{u}; \mathbf{W}^{out}))$, i.e., $\mathcal{F}({y(\mathbf{u}; \mathbf{W}^{out}))})$, where:
\begin{equation}
   p( h \mid y(\mathbf{u}; \mathbf{W}^{out})) =
\mathcal{N} (\mathcal{F}({y(\mathbf{u}; \mathbf{W}^{out}))}), \epsilon_f ^2),
\label{esn:eq:gaussianlikelihood}
\end{equation}
and write the joint probability as the product of two Gaussians by assuming conditional independence:
\begin{equation}
  p( \Tilde y, h \mid y(\mathbf{u}; \mathbf{W}^{out}))
     =
\mathcal{N} (y(\mathbf{u}; \mathbf{W}^{out}), \epsilon_d ^2) 
\cdot 
\mathcal{N} (\mathcal{F}({y(\mathbf{u}; \mathbf{W}^{out}))}), \epsilon_f ^2),
    \label{esn:eq:gaussianlikelihood2}
\end{equation}
%

When applying the negative log-likelihood, we get a sum of two loss terms and the uncertainty parameter as the total loss $L$:
\begin{equation} \label{esn:eq:gaussianlikelihood_2}
\begin{aligned}
-\log p( \Tilde{y}, h \mid y(\mathbf{u}; \mathbf{W}^{out}) 
& \propto  
\frac{1}{{2{\epsilon^2 _d}}} J_{data}(\mathbf{W}^{out}; N_t)
+
\frac{1}{{2{\epsilon^2 _f}}} J_{physics}
(\mathbf{W}^{out}; N_f)
 +  \log {\epsilon_d \epsilon_f}  \\
  & = 
L( \mathbf{W}^{out}, \mathbf{\epsilon}; \mathbf{N})
\end{aligned}
\end{equation}
where $\epsilon = \{\epsilon_d, \epsilon_f\}$ denotes the adaptive parameters that balance the contributions of both data loss and physics-informed loss terms, and $\mathbf{N} = \{N_t, N_f\}$. 
We can observe that the role of $\frac{1}{{2{\epsilon^2 _d}}}$ is that of $\lambda_{data}$, whereas $\frac{1}{{2{\epsilon^2 _f}}}$ functions as $\lambda_{physics}$. 
Notice that the total loss $L$ is proportional to the negative log-likelihood since each loss term is not evaluated for the whole $\mathbf{u}$, but applied to their respective dataset: $J_{data}$ to the $N_t$ data points, and $J_{physics}$ to the $N_f$ collocation points. 
Besides, the target (true value) for the second Gaussian is zero since collocation points do not have labels and $\mathcal F(\mathbf{y}) \equiv 0 $, which means we want the network output to respect the physics laws. 

Now, the total loss $L( \mathbf{W}^{out}, \mathbf{\epsilon}; \mathbf{N})$ is minimized with respect to $\mathbf{W}^{out}$ and also $\mathbf{\epsilon}$ using gradient descent or L-BFGS to update the parameters. A low $\epsilon_f$ results in higher punishment for the physics-informed loss $J_{physics}$ term. 
On the other hand, $\epsilon_f$ will not increase indefinitely due to the term $\log { \epsilon_f} $.

If $\epsilon$ is negative,  $\log { \epsilon_f} $ is not defined, which can cause problems during the training process.
To deal with this situation, we change variables by defining
$\mathbf{s} = \log(\epsilon^2)$, and rewriting the loss as \cite{Xiang2022}:
\begin{multline}
L( \mathbf{W}^{out}, \mathbf{s}; \mathbf{N}) = \frac{1}{{2}} \exp{(-s_d)}  J_{data}(\mathbf{W}^{out}; N_t) \\ + 
\frac{1}{{2}} \exp{(-s_f)}  J_{physics} (\mathbf{W}^{out}; N_f) 
+ s_d + s_f, 
\label{esn:eq:expadaptative}
\end{multline}
where $\mathbf{s} = \{s_d, s_f\}$. This exponential mapping allows $\mathbf{s}$ to be adapted in an unconstrained way, facilitating the optimization of the loss.
The final pseudocode for training the PI-ESN with external Inputs and self-Adaptive balancing loss is given in Algorithm \ref{algor:adaptive_PI_ESN}.


\begin{algorithm2e}
\SetKwInput{KwIn}{input}
\SetKwInput{KwOut}{output}
\KwIn{
$M$, $K$, $\mathbf{s}=[s_d, s_f]^T$,
$\mathcal{F}$, 
 \{$\mathbf{u}[n]: n = 1,\dots, N_t+N_f$\},
 \{$\mathbf{\hat y}[n]: n = 1,\dots, N_t$\};
}

Using $\{ (\mathbf{u}[n], \mathbf{\hat y}[n]):n = 1,\dots, N_t\}$, build $\mathbf{X}_t$  by \autoref{eq:Xreservoir} and $\mathbf{\hat Y}$ by \autoref{eq:Ytarget}; \tcp{training data}


Pretrain \eric{ESN's output} weights $\mathbf{W}^{out}$ by ridge regression w/ \autoref{eq:ridgeregression} using  $\mathbf{X}_t$ and  $\mathbf{\hat Y}$;

\For{$M$ \emph{iterations}}{

Generate $\mathbf{X}_f$ 
using \autoref{eq:stateupESN},  \autoref{eq:outputESN}, 
\eric{current} $\mathbf{W}^{out}$, 
and $\mathbf{u}[n]$ for timesteps $n = N_t + 1,\dots, N_t + N_f$; \tcp{collocation points}

\For{$K$ \emph{iterations}}{
  \tcp{Adapting $\mathbf{W}^{out}, s_d, s_f$ to minimize total loss}
  Compute ESN's outputs for the data points $\mathbf{Y}_t = \mathbf{W}^{out} \mathbf{X}_t$ and for collocation points $\mathbf{Y}_f = \mathbf{W}^{out} \mathbf{X}_f$
  
  Define $J_\mathrm{data}(\mathbf{W}^{out})$ loss on $\mathbf{Y}_{t}$ and target $\mathbf{\hat Y}$ as in \autoref{eq:Jdata};
 
 Define $J_\mathrm{physics}(\mathbf{W}^{out})$ loss on $\mathcal{F}(\mathbf Y_f)$ as in \autoref{esn:eq:custofisico}; \\
 Combine both losses into 
 $L(\mathbf{W}^{out}; \mathbf{s})$ as in \autoref{esn:eq:expadaptative} 
 and compute its gradient with respect to 
  $\mathbf{W}^{out}, s_d, s_f$;\\
  Update $\mathbf{W}^{out}, s_d, s_f$ with an optimizer and the obtained gradients;}}
\KwOut{$\mathbf{W}^{out}$}
\caption{Training of PI-ESN with external Inputs and Self-Adaptive Balancing Loss (PI-ESN-a).}
\label{algor:adaptive_PI_ESN}
\end{algorithm2e}

\newpage

\section{Experiments}
\label{sec:experiments}
This section applies the proposed PI-ESN-a with external control inputs and adaptive balanced loss to modeling the Van der Pol Oscillator dynamical system \cite{Hafeez2015vdp},
the four-tank MIMO (multiple-input, multiple-output) system 
\eric{and the electric submersible pump (ESP). In addition, a MPC experiment is presented with the four-tank system to track the level of two tanks.
}

\subsection{Van der Pol Oscillator}

\subsubsection{Model}

\begin{figure}[h!]
    \centering
    \includegraphics[width = 0.9\textwidth]{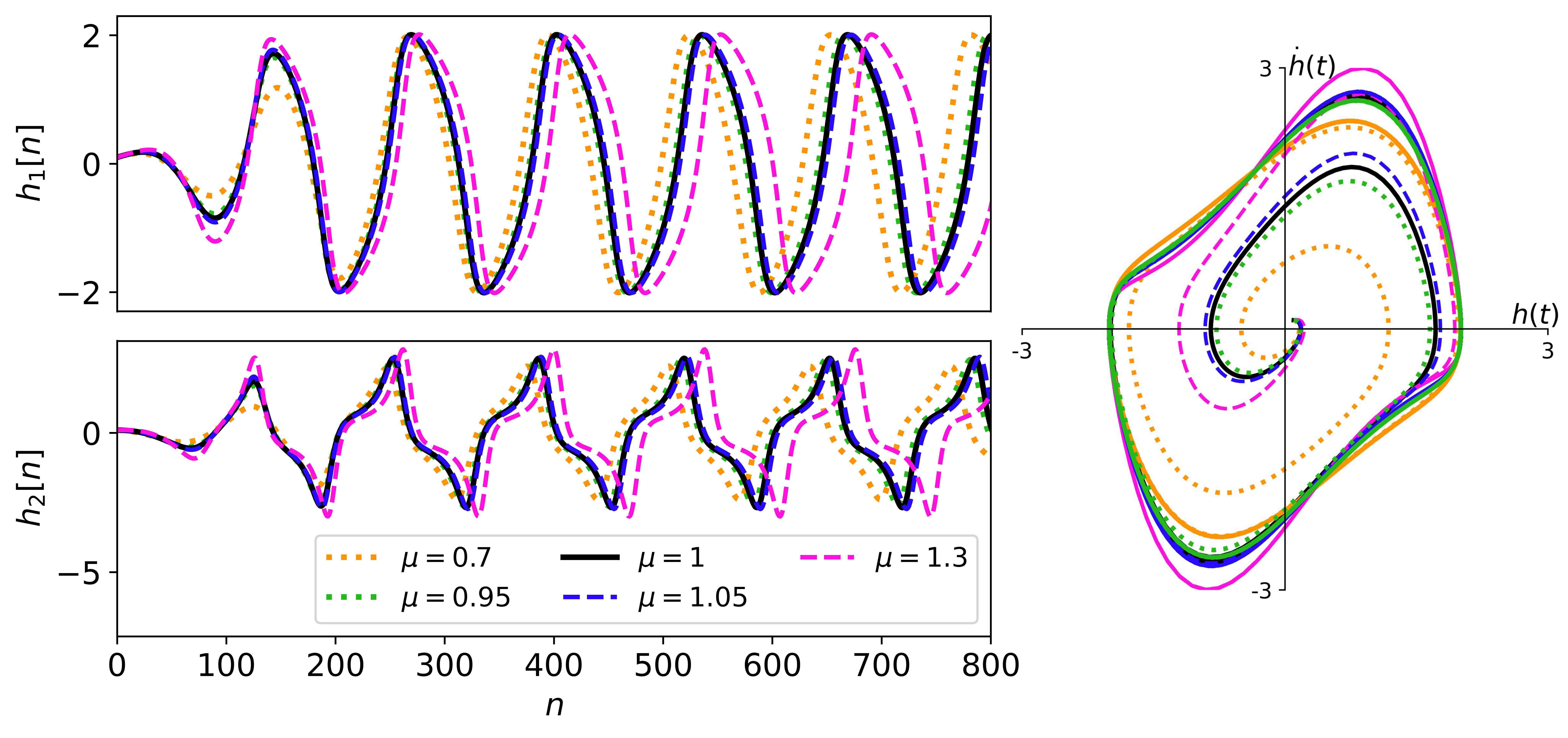}
    \caption{Van der Pol oscillator system with initial conditions $h_1(0) = h_2(0) = 0.1$ shows the variation of oscillation depending on the damping parameter ($\mu$). 
    }
    \label{fig:vanderpol}
\end{figure}

Extensive research has been conducted on the Van der Pol Oscillator, 
aiming at enhancing the approximations of solutions to non-linear systems. 
It is a self-oscillatory dynamical system widely recognized as a valuable mathematical model that can be employed in more complex systems. 
Its second-order ordinary differential equation featuring cubic nonlinearity \cite{Tsatsos2008} is described as follows:
\begin{equation}
\ddot h - \mu (1- h^2)\dot h + h = 0,
    \label{esn:eq:vanderpol_2order}
\end{equation}
where \textcolor{blue}{$h(t)$ is a function of time with the position coordinate} and $\mu$ represents the damping parameter, which influences the system's oscillation as shown in \autoref{fig:vanderpol}. 

\autoref{esn:eq:vanderpol_2order} can be written in two-dimension form, with $u$ as the control input:
\begin{equation}
\begin{array}{l}
{{\dot h}_1} = {h_2}\\
{{\dot h}_2} = \mu (1 - {h_1}^2){h_2} - {h_1} + u
\end{array}
    \label{esn:eq:vanderpol_2dimension}
\end{equation}

\subsubsection{Dataset}
\begin{figure}[h!]
    \centering
    \includegraphics[width = 0.9 \textwidth]{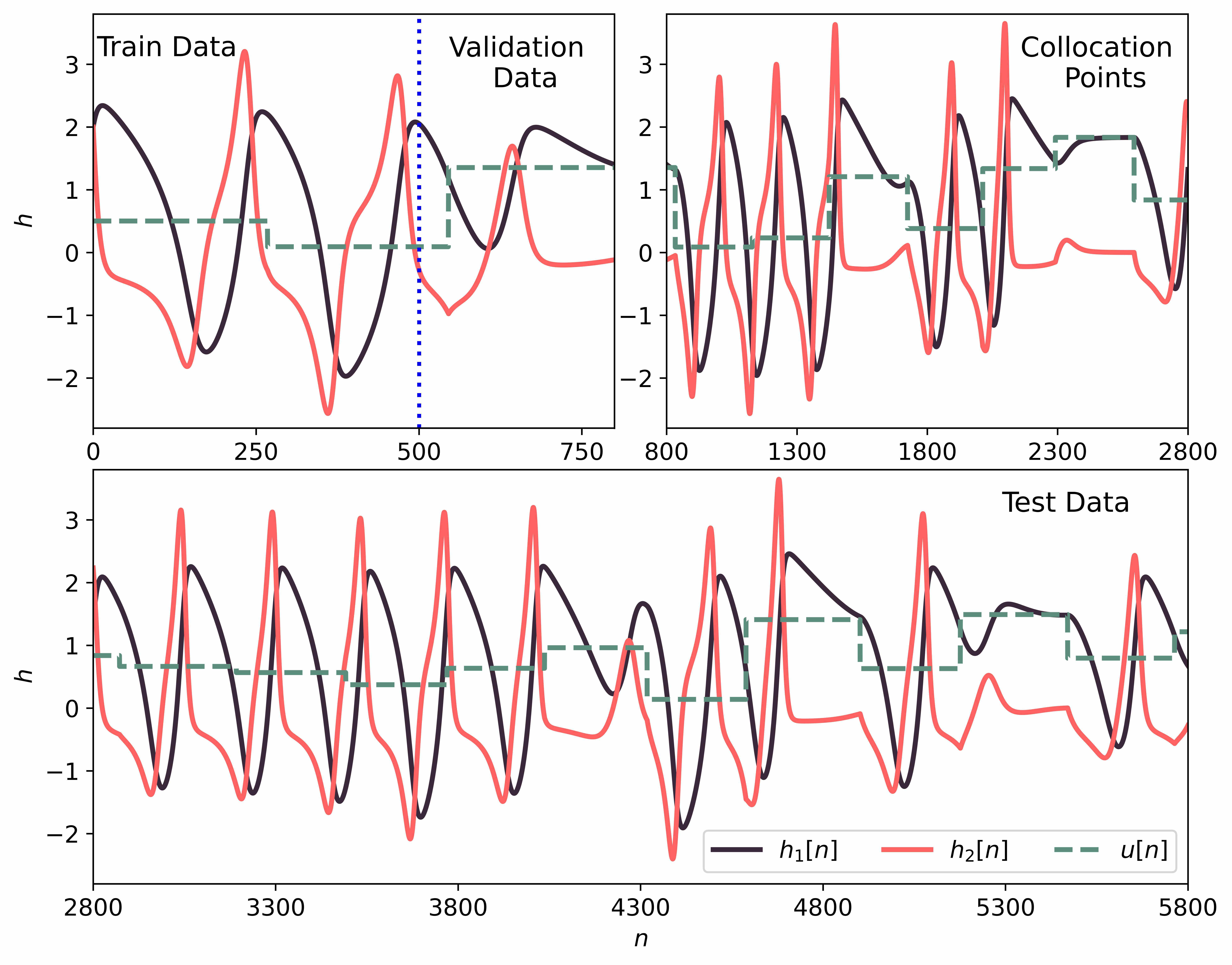}
    \caption{An example of one of the simulated Van der Pol system is depicted in three plots. The first plot displays the training set $N_{te}$ and the validation set $N_{ve}$, separated by a dashed blue line used for hyperparameter tuning during the training of the ESN. The first plot corresponds to the total number of points in the training set $N_t = 800$. The second plot represents the region where physics-informed training is performed using 2000 collocation points. In this region, only the random input values $u$ at the $N_f$ collocation points are utilized for regularization purposes, while the labels $h_1$ and $h_2$ are not used. Lastly, the third plot illustrates the test dataset consisting of 3000 points, which is utilized for the analysis of the PI-ESN-a.
    }
    \label{fig:vanderpol_simulated_system}
\end{figure}

The dataset to train the ESNs considered in the following section 
was generated using \autoref{esn:eq:vanderpol_2dimension},
which was simulated by an explicit Euler method with a time step $dt = 0.03$ sec, initial values of $h_1 = h_2 = 2$, and $\mu = 1$. 
The input signal $u[n]$ was generated using an Amplitude Modulated Pseudo-Random Bit Sequence (APRBS).
An illustration of how the dataset is organized is shown
\autoref{fig:vanderpol_simulated_system}, where the simulated signal is contiguously split into training data, validation data, collocation points and test data, in this order, and with
500, 300, 2000, and 3000 timesteps, respectively, unless otherwise stated.
The total unlabeled dataset consists of the concatenation of the collocation points with the test data, totaling 5000 timesteps. 

\subsubsection{ESN settings}
\label{sec:vdp:settings}
For the experiments reported below, the \eric{ESN} was trained in two stages:
1) first pre-training an ESN by conventional ridge regression and optimizing its main hyperparameters; and 2) subsequently refining this ESN by applying the proposed method, as described in Algorithm \ref{algor:adaptive_PI_ESN}.
For all experiments, unless otherwise stated, the reservoir size is $N_x = 100$, the leak rate is $\alpha = 1$, and the spectral radius is $\rho (\mathbf{W}) = 0.8$.
 
In the first stage, hyperparameter optimization was performed using a training set ($N_{te}=500$) and a validation set ($N_{ve}=300$). A grid search was executed for input scaling ($\delta_{in}$) and feedback scaling ($\delta_{fb}$) within the range of 0.05 to 0.95, with increments of 0.05. 
Additionally, the Tikhonov regularization factor $\gamma$ was explored over magnitudes ranging from $10^{-2}$ to $10^{-7}$, using a resolution of $10^{-1}$. With the optimal resulting parameter values, the ESN was retrained using a combined training and validation set of size $N_{t} = N_{te} + N_{ve} = 800$. 
In the second stage, the resulting ESN was then used as the initial guess for the \eric{PI-ESN-a}, which was optimized using the L-BFGS algorithm from the TensorFlow framework.

\subsubsection{\eric{PI-ESN-a} improves over ESN}
\label{sec:piesn_improves}
The first experiment consisted of validating the proposed \eric{PI-ESN-a} in a limited training data scenario.
%
%
\autoref{tab:vdp_mse} shows the MSE for the collocation points and the test set,
for three ESN architectures: 
conventional ESN, 
PI-ESN-i (``i'' for \textit{with external inputs}),
and PI-ESN-a.
For each case, the performance was averaged
over five different randomly generated ESNs and input values.
For the PI-ESN-i, $\lambda_{data}=\lambda_{phy}=1$.
The PI-ESN-a was initialized with $s_d = s_f = 1$. 
Notice that both PI-ESNs use $N_f = 2000$ collocation points for physics-informed training, while the ESN \eric{uses only} labeled training data.
%
%

\begin{table}[htb!]
\centering
\begin{tabular}{ccc}
\hline
\multicolumn{1}{c}{Architecture} & Collocation Points & Test set \\ \hline
ESN                              & 1.5217             & 1.5485 \\
PI-ESN-i                           & 0.1967             & 0.2575 \\
PI-ESN-a                  & \textbf{0.1161}             & \textbf{0.1912}  \\ \hline
\end{tabular}
\caption{\label{tab:vdp_mse} Results for the Forced Van der Pol Oscillator: Average MSE of five runs for ESN and PI-ESN architectures with external inputs.}
\end{table}
The results show that the additional physics-informed training of ESNs in limited data scenarios significantly reduces the average error on the collocation points set and on the test set.
The PI-ESN-i, and the PI-ESN-a achieve an average reduction of $84.8\%$ and $89.5\%$, respectively, over the conventional ESN.  
The PI-ESN-a improved $30.9\%$ over the PI-ESN-i.
\eric{Note that standard PI-ESN is not part of the experiments because it does not handle control inputs.
}

\subsubsection{PI-ESN-a generalizes better for unseen control inputs}
\label{sec:piesn_generalizes}

\begin{figure}[htb!]
    \centering    
    \includegraphics[width = \textwidth]{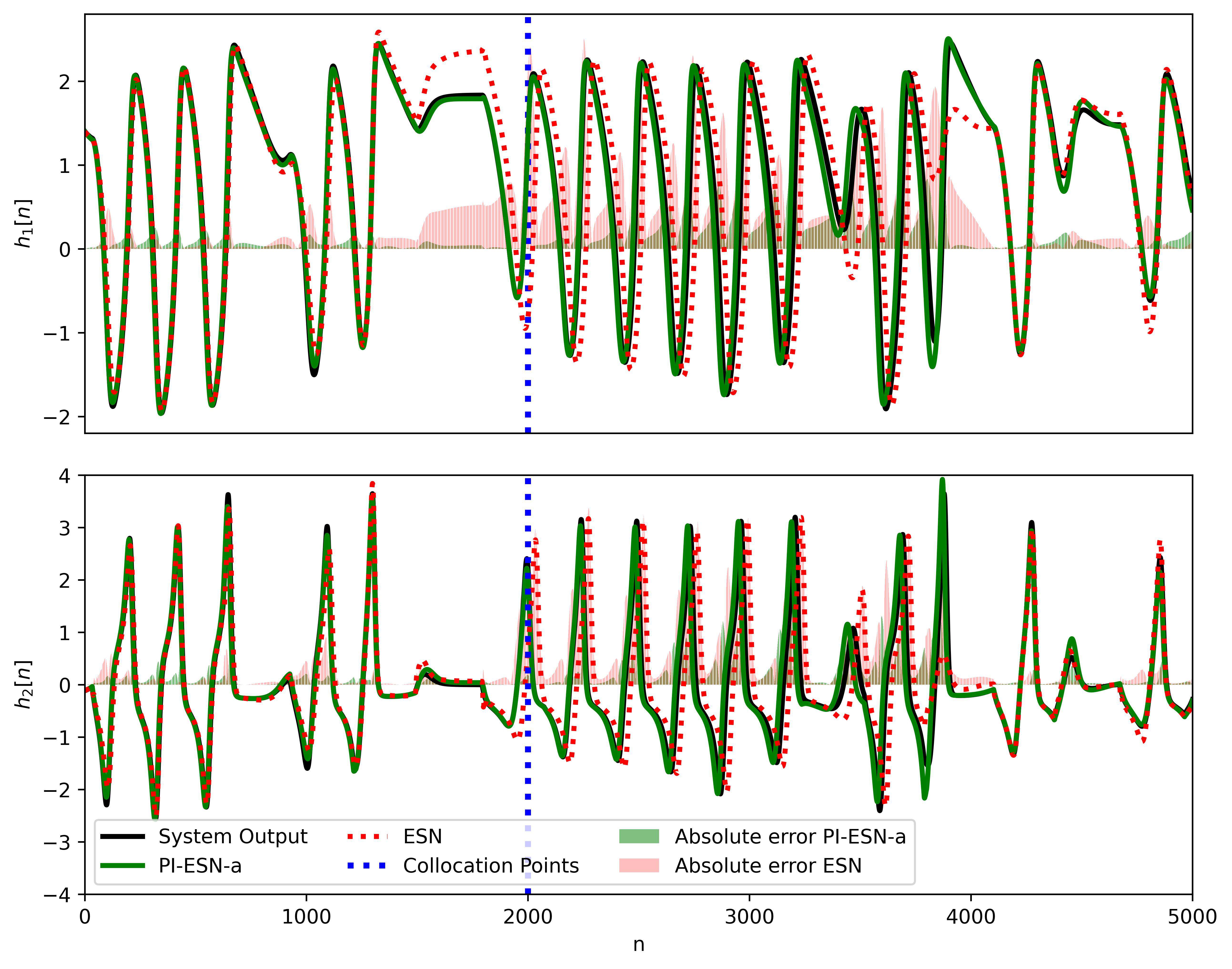}
        
    \caption{Prediction of the adaptive PI-ESN-a for Van der Pol oscillator after training. The blue dashed vertical line splits the region between collocation points (left) and test set (right).
    In the background, it is possible to observe the absolute error of the ESN and PI-ESN-a with the actual system output. This prediction refers to the collocation points and test data of the system presented in \autoref{fig:vanderpol_simulated_system}. The MSE for the collocation points region was found to be 0.1204 for the ESN and 0.0101 for the PI-ESN-a. In the test region, the corresponding MSE values were 0.8601 and 0.1741, respectively.
    }
    \label{fig:piesn_oscilador}
\end{figure}

By looking at the prediction results more closely of a PI-ESN-a in \autoref{fig:piesn_oscilador}, it is possible to verify that it can refine the prediction performance of a conventional ESN for unseen control inputs.
Here, the reservoir size is changed to $N_x = 200$.
The first 2000 timesteps refer to the collocation points, while the remaining 3000 timesteps form the test dataset of the system presented in \autoref{fig:vanderpol_simulated_system}. 
The MSE of the ESN and PI-ESN-a for the collocation points were 0.1204 and 0.0101, respectively. 
In the test set, the corresponding MSE values were 0.8601 and 0.1741, respectively.

\begin{figure}[htb!]
    \centering
    \includegraphics[width = \textwidth]{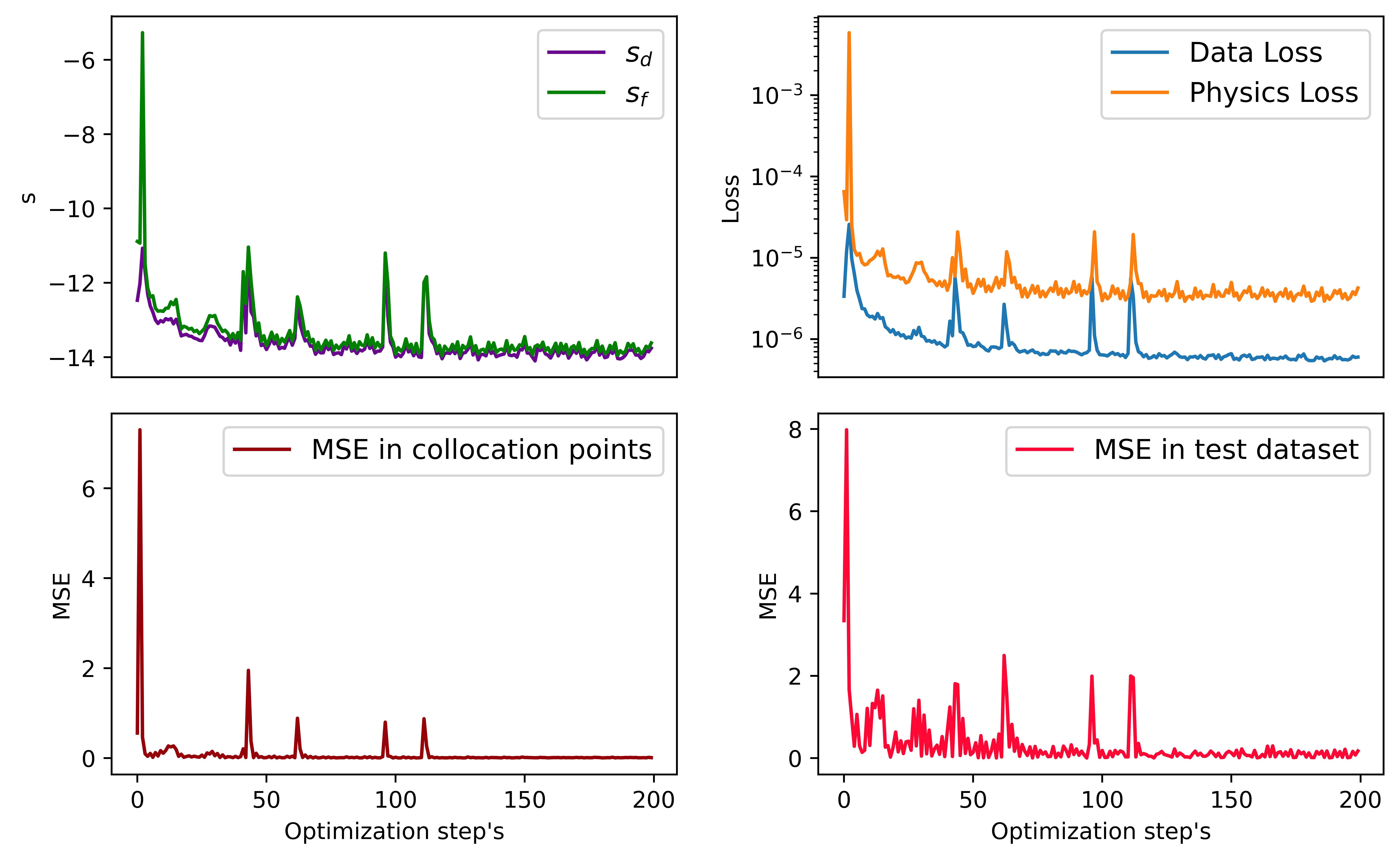}
    \caption{Evolution of the adaptive weights ($s_d, s_f$), the losses functions ($J_{data}, J_{phy}$) and the MSE during the physics training of the PI-ESN-a. 
    }
    \label{fig:adaptive_vanderpol}
\end{figure}

The corresponding PI-ESN-a training process  is shown in \autoref{fig:adaptive_vanderpol}, where
the adaptive loss parameters $s_f$ and $s_d$,
the data loss and physics-informed loss values, the MSE for the collocation points, and the MSE for the test dataset are displayed as training evolves. 
The $J_{data}$ and $J_{physics}$ values at the end of the experiment are $6 \times 10^{-7}$ and $4.25 \times 10^{-6}$, respectively. Moreover, the final values of the $[s_d,s_f]$ parameters are $[-13.75, -13.61]$.  Analyzing the graphs, it is evident that there was a reduction in both physics and data loss functions, as well as errors in the test dataset and collocation points.

The results demonstrate that training with physics laws can refine and improve the prediction performance of the ESN for unseen control inputs that are not available during training.
The experiments confirm the generalization and regularization capacities of physics-informed training.

\subsubsection{Effect of reservoir size and labeled data size}

Here, we investigate how PI-ESN-a compares to ESN for different reservoir sizes and labeled dataset sizes.
The hyperparameters' values are as those used in Section~\ref{sec:piesn_improves}.
\autoref{tab:Reservoirexperiment} shows the MSE for reservoir sizes from 100 to 400 on the set of collocation points and on the test set, where each result was averaged over 6 randomly generated inputs and 5 randomly generated ESNs, thus amounting to 30 runs for each reservoir size and 120 runs for the whole experiment since four different reservoir sizes are tested.
Numerical instability has taken place for 22 runs, and, for this reason, they were excluded from the analysis.
The results in the table reveal that while the average MSE is always considerably reduced when going from the ESN to the extra training of PI-ESN-a, it does not approach zero in either of the configurations. This observation can be attributed to the inherent error introduced by the explicit Euler approximation utilized in the physics-informed training process, as emphasized by \cite{Racca2021}.


\begin{table}[htb!]
\centering
\begin{tabular}{ccccc}
\hline
\multicolumn{1}{c}{\multirow{2}{*}{\begin{tabular}[c]{@{}c@{}}Reservoir\\  Size\end{tabular}}} & \multicolumn{2}{c}{\makecell{Collocation Points}  } & \multicolumn{2}{c}{Test set}                \\ \cline{2-5} 
\multicolumn{1}{c}{}                                                                           & ESN               & \makecell{PI-ESN-a}             & ESN         & \makecell{PI-ESN-a}      \\ \hline
100    & 1.255             & 0.346              & 1.758       & 0.435                     \\
200                                                                                            & 0.873             & 0.400              & 1.230       & 0.481        \\
300                                                                                            & \textbf{0.514}             & \textbf{0.274}              & \textbf{0.815}       & \textbf{0.346 }                    \\
400                                                                                            & 1.373             & 0.450              & 1.762       & 0.614 \\ \hline
\end{tabular}
\caption{\label{tab:Reservoirexperiment} Average MSE for the conventional ESN and the PI-ESN-a as a function of the reservoir size ($N_x$) for the Van der Pol oscillator. Each value is obtained by averaging the MSE of around $30$ experiments.}
\end{table}

\begin{figure}[h!]
    \centering
    \includegraphics[width = 0.8\textwidth]{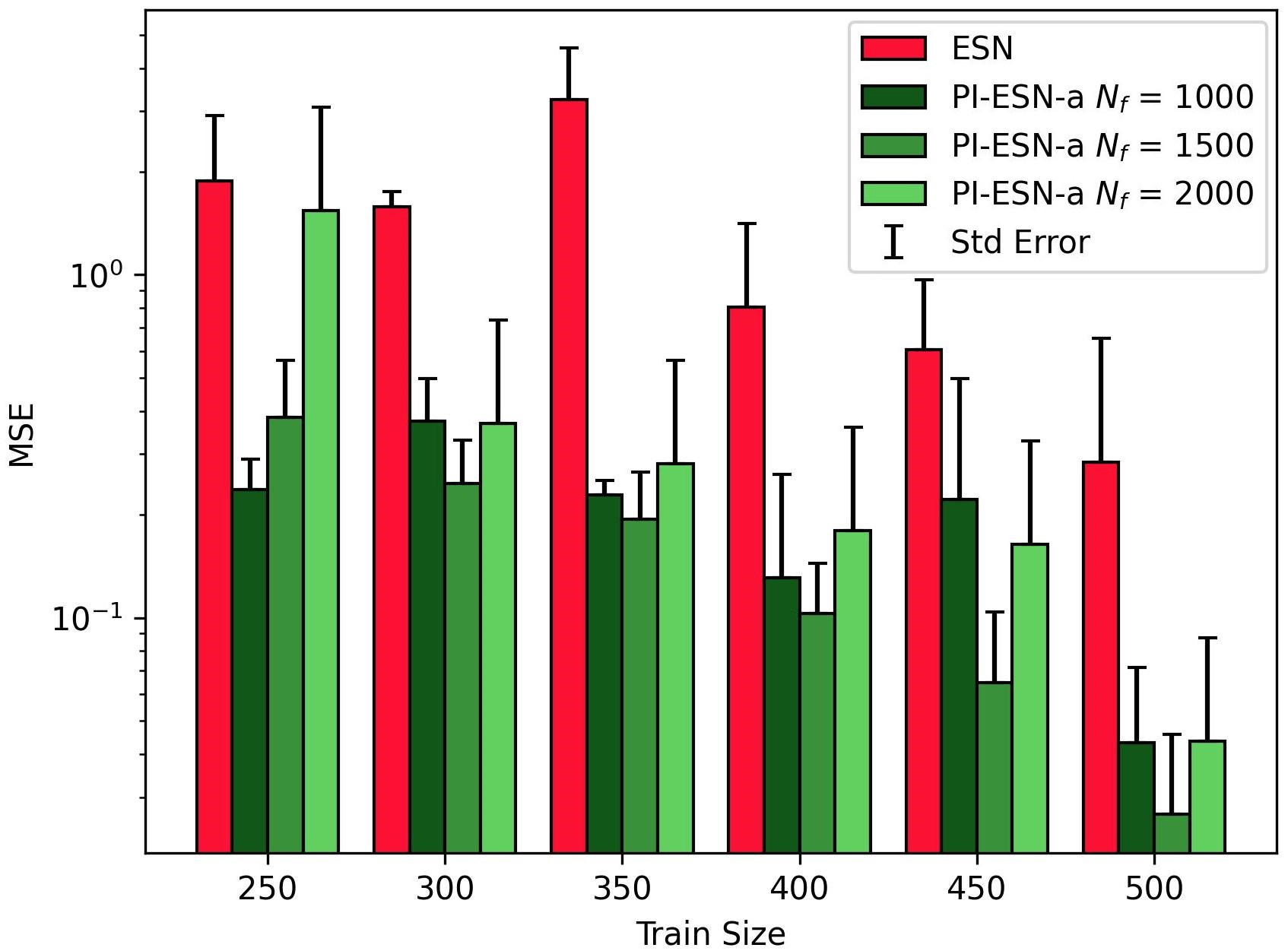}
    \caption{ 
    The average MSE of the PI-ESN-a is displayed as a function of the size $N_t$ of the labeled training dataset for three different cases of $N_f$ --- $1000$, $1500$, and $2000$ points --- representing the number of collocation points. The presentation is based on six runs with diverse random seeds that influence the initial weights of the ESN, while keeping the control signal constant. Furthermore, the results encompass the MSE of ESN for comparative analysis.
    }
    \label{fig:pi_variation}
\end{figure}

The more data for training are available, the better a traditional ESN will perform. In
\autoref{fig:pi_variation}, the average MSE for six randomly generated reservoirs are shown for various training set sizes ($N_t$) and for different numbers of collocation points ($N_f= [1000, 1500, 2000]$) used in the physics-informed training. 
The evaluation is performed over 2100 time steps consisting of collocation points and test data.
 It is noticeable that more training data improves both ESN and PI-ESN-a. The latter always yields a regularization effect in the limited data scenario, reducing the error of the respective pre-trained ESN. 

\subsubsection{PI-ESN-a's robustness to parameter model uncertainty}

In real-world scenarios, it is expected to observe disparities between the actual system dynamics and its model. 
To evaluate the proposed PI-ESN-a's robustness to uncertainties in the model equations, the parameter $\mu$, which directly impacts the oscillation behavior (\autoref{fig:vanderpol}), was artificially perturbed. The resulting perturbed model was subsequently used in the physics-informed loss for training the PI-ESN-a, while the data loss was still based on the labeled system data kept at a reference value of $\mu=1$.
The hyperparameters' values and other settings are the same ones used in
Section~\ref{sec:piesn_generalizes}.
\begin{figure}[h!]
    \centering
    \includegraphics[width = \textwidth]{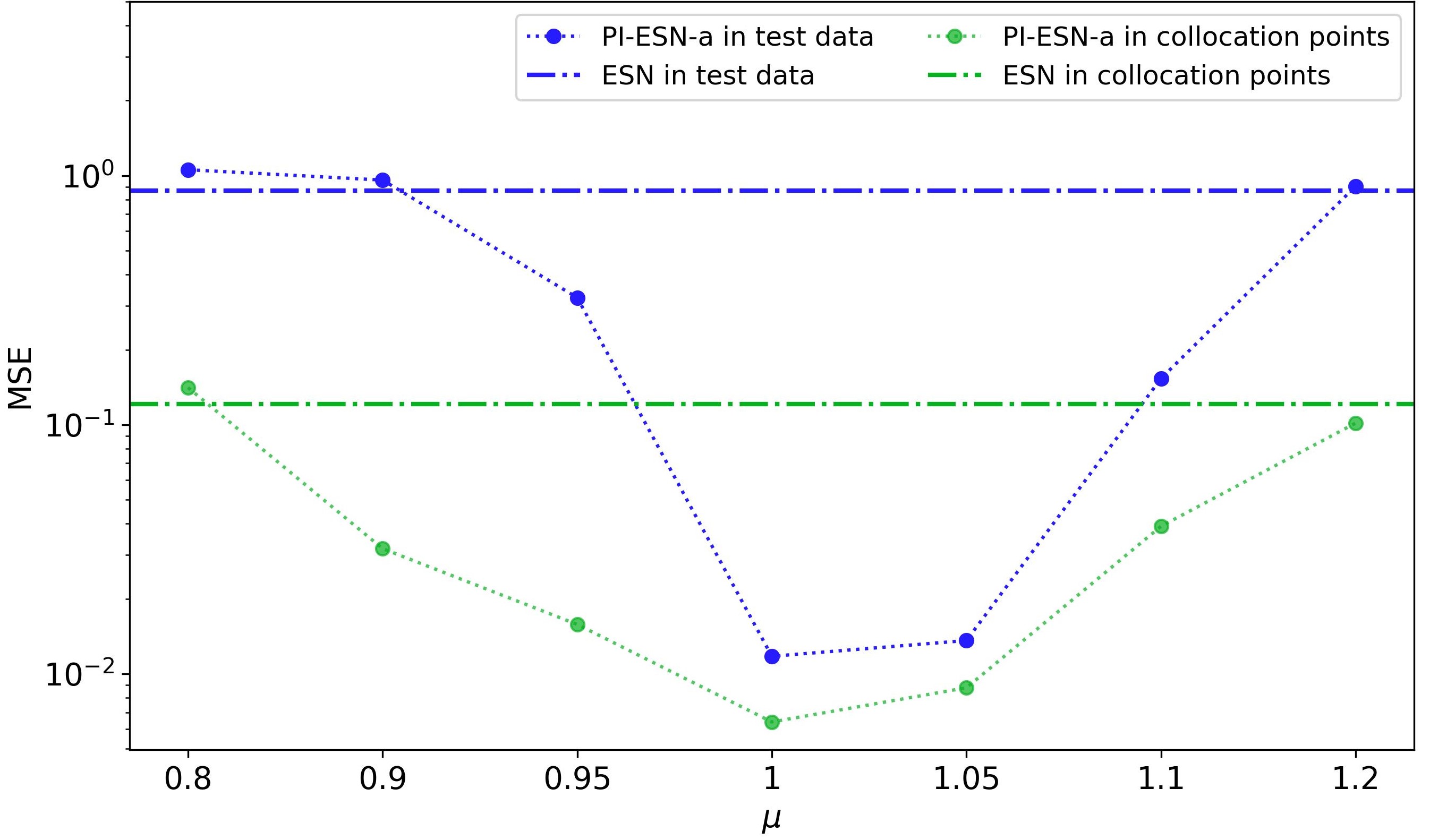}
    \caption{
    Disturbing damping parameter ($\mu$) from \autoref{esn:eq:vanderpol_2dimension} used in the physics-based loss for training the PI-ESN-a.
    The MSE is displayed for the collocation points region, test set, and total dataset (as shown in the data split in \autoref{fig:vanderpol_simulated_system}) for the PI-ESN-a. The horizontal lines represent the constant MSE values of the ESN for the collocation points region and test set, as the damping parameter alteration only affects the calculation of the physics function (\autoref{eq:euler}).}
    \label{fig:incerteza_variation}
\end{figure}

\autoref{fig:incerteza_variation} shows the MSE for a PI-ESN-a trained considering different disturbances in $\mu$. 
When a disturbance is too strong, i.e., $\mu$ exceeds 1.1 or falls below 0.95, the test error (blue dots) of PI-ESN-a is slightly worse than that of ESN (horizontal blue line). 
However, for disturbances resulting in $\mu \in [0.95,1.1]$, the proposed PI-ESN-a continues to improve and regularize ESN's prediction, as seen by the dots below their respective horizontal lines of the same color. 
Consequently, this outcome highlights the capability of the PI-ESN-a to achieve lower error rates than the ESN, despite the presence of a parametric error associated with the $\mu$ parameter in the physics equation, as well as errors arising from the derivative calculated using the explicit Euler method.

\subsection{Four-tank system}

\subsubsection{Model}
The four-tank system consists of interconnected tanks with two pumps that can be used to control the flow rate into the tanks. Usually, the levels of the tanks are controlled by manipulating the voltages applied to pumps \cite{ALVARADO2006, Johansson2000}. 
Accurate identification of system dynamics is important when designing predictive controllers. 
The four-tank process is characterized by the following system of differential equations:
\begin{equation}
\begin{aligned}
\frac{d}{dt}{h_1}(t) &=  - \frac{{{a_1}}}{{{A_1}}}\sqrt {2g{h_1}(t)}  + \frac{{{a_3}}}{{{A_1}}}\sqrt {2g{h_3}(t)}  + \frac{{{\gamma _1}{k_1}}}{{{A_1}}}{V_1}(t)\\
\frac{d}{dt}{h_2}(t) &=  - \frac{{{a_2}}}{{{A_2}}}\sqrt {2g{h_2}(t)}  + \frac{{{a_4}}}{{{A_2}}}\sqrt {2g{h_4}(t)}  + \frac{{{\gamma _2}{k_2}}}{{{A_2}}}{V_2}(t)\\
\frac{d}{dt}{h_3}(t) &=  - \frac{{{a_3}}}{{{A_3}}}\sqrt {2g{h_3}(t)}  + \frac{{(1 - {\gamma _2}){k_2}}}{{{A_3}}}{V_2}(t)\\
\frac{d}{dt}{h_4}(t) &=  - \frac{{{a_4}}}{{{A_4}}}\sqrt {2g{h_4}(t)}  + \frac{{(1 - {\gamma _1}){k_1}}}{{{A_4}}}{V_1}(t),
\end{aligned}
    \label{esn:eq:fourtankequations}
\end{equation}
where: $h_i$ denotes the level of each tank $i$; and $V_1$ and $V_2$ represent the voltage applied to the pumps. 
The cross-sectional area of each tank and the cross-sectional area of the bottom orifice are represented by $A_i$ and $a_i$, respectively. The constants $k_1$ and $k_2$ relate the flow rate to the applied voltage in the pump, while the valves' openings are denoted by $\gamma_1$ and $\gamma_2$. 
The corresponding values for each parameter, along with their units, are presented in \autoref{tab:unitsfourtanks}.
\begin{figure}[h!]
    \centering
    \includegraphics[width = 0.8\textwidth]{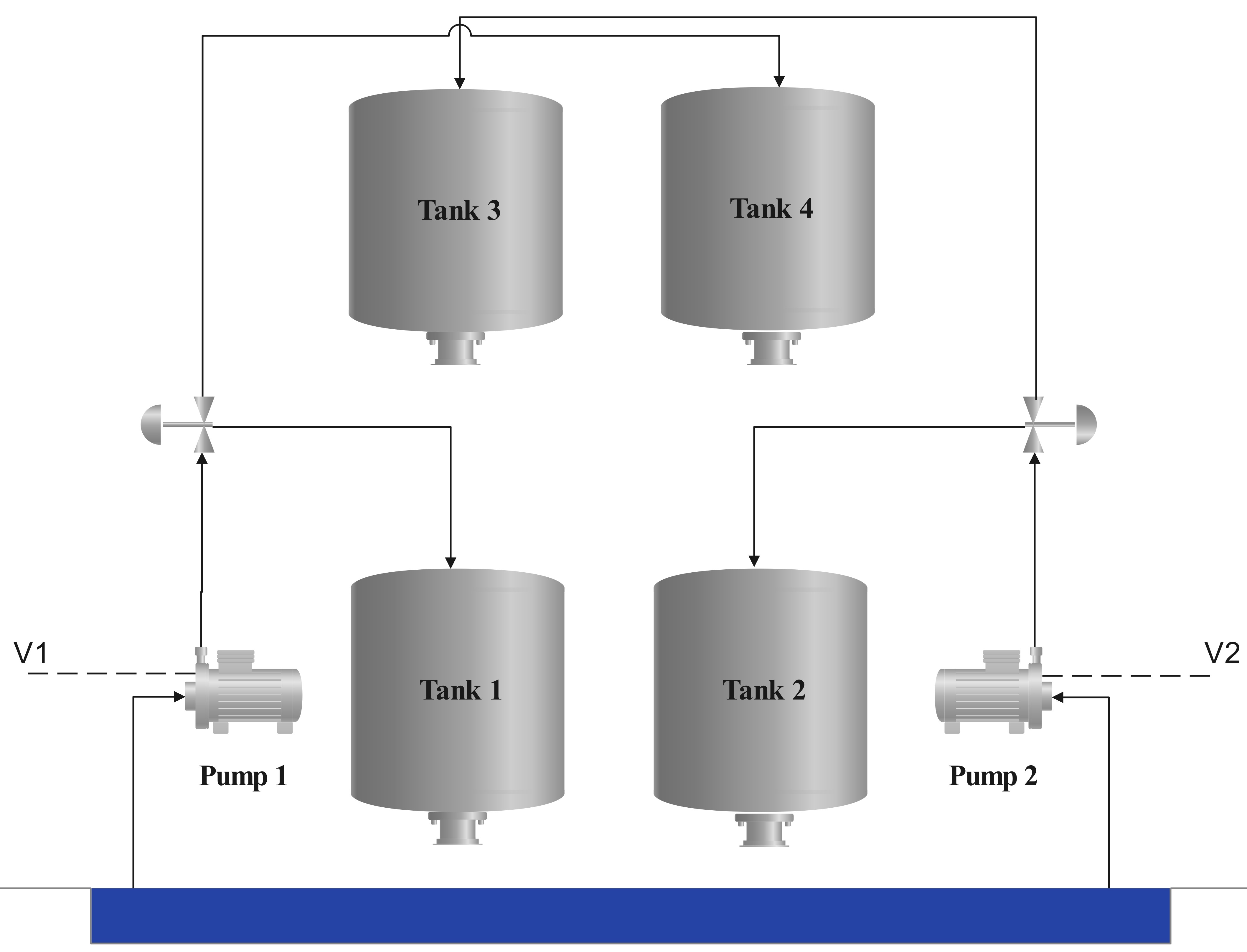}
    \caption{Nonlinear process of control the water levels in a four-tank system.}
    \label{fig:fourtanks}
\end{figure}

\begin{table}[h!]
\centering
\begin{tabular}{lll}
\hline
Parameter & Value & Unit        \\ \hline
$A_1, A_3$    & 28    & $cm^2$         \\
$A_2, A_4$    & 32    & $cm^2$         \\
$a_1, a_3$    & 0.071 & $cm^2$         \\
$a_2, a_4$    & 0.071 & $cm^2$         \\
g         & 981   & $cm^2$.$s^{-2}$     \\
$k_1, k_2$    & 1     & $cm^3$.$V^{-1}$.$s^{-1}$ \\
$\gamma_1$    & 0.7   &             \\
$\gamma_2$    & 0.6   &             \\ \hline
\end{tabular}
\caption{\label{tab:unitsfourtanks} Model parameters of the four-tank system.}
\end{table}

\subsubsection{Dataset}
\begin{figure}[h!]
    \centering
    \includegraphics[width = 0.85\textwidth]{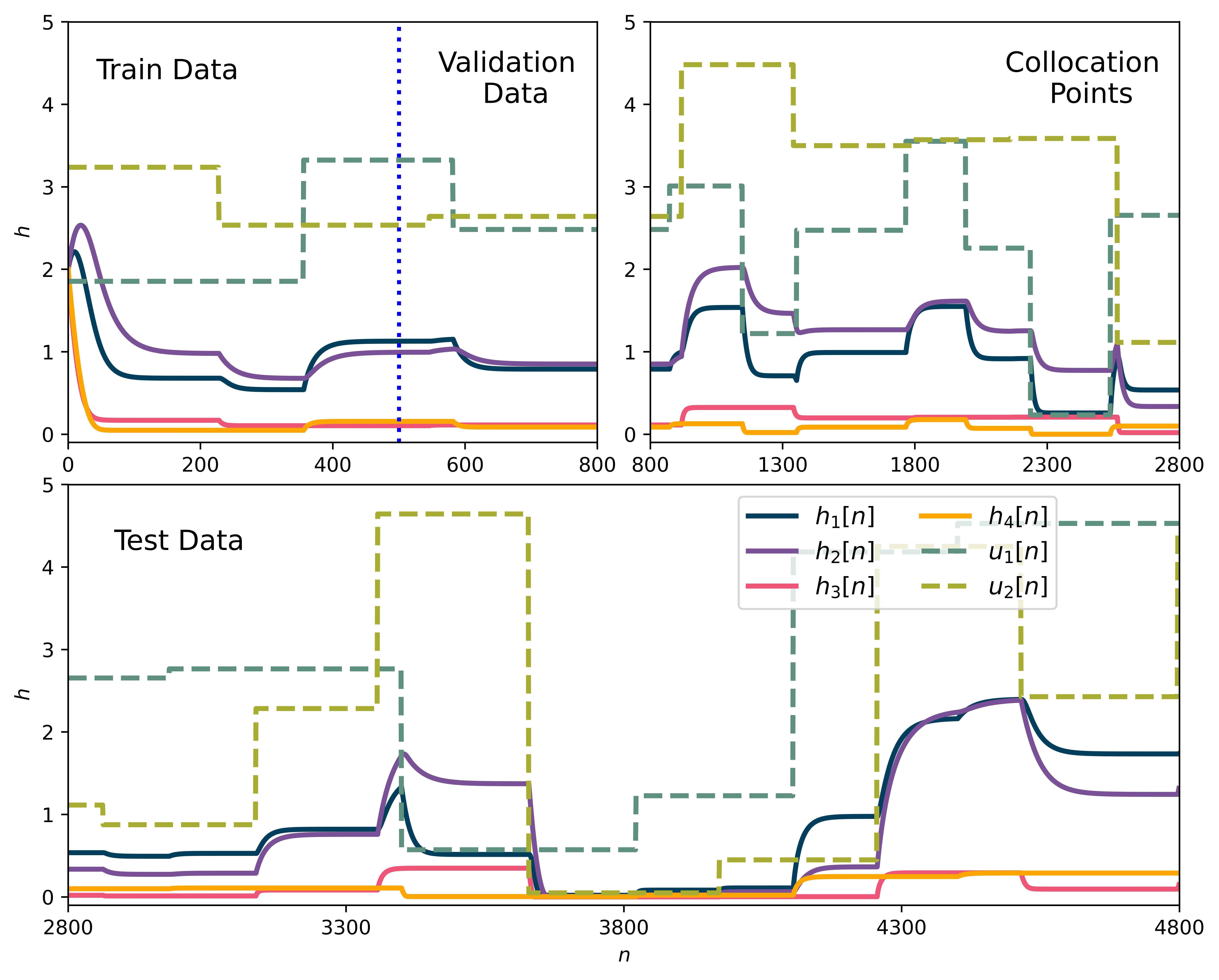}
    \caption{An example of one of the simulated four-tank systems is depicted in three plots. The first plot displays the training set $N_{te}$ and the validation set $N_{ve}$, separated by a dashed blue line used for hyperparameter tuning during the training of the ESN. The first plot corresponds to the total number of points in the training set $N_t = 800$.       The second plot represents the region where physics-informed training is performed using 2000 collocation points. In this region, only the random input values $u$ at the $N_f$ collocation points are utilized for regularization purposes, while the labels $h_1$, $h_2$, $h_3$, and $h_4$ are not used. Lastly, the third plot illustrates the test dataset consisting of 2000 points, which is utilized for the analysis of the PI-ESN-a.}
     \label{fig:fourtanks_sys}
\end{figure}

The four-tank process in \autoref{esn:eq:fourtankequations} was simulated using an explicit Euler method with a time step  of $dt = 1$s, and initial values $h_1 = h_2 =h_3 = h_4 = 2$. 
The input signal was generated using a pseudo-random binary sequence (PRBS), for both $V_1$ and $V_2$. 
The generated dataset, shown in \autoref{fig:fourtanks_sys}, is split into 
training data ($N_{t}=800$),
collocation points ($N_{f}=2000$),
and test data ($N_{test}=2000$).

\subsubsection{ESN settings}

The network settings were set as follows unless otherwise stated.
The reservoir size, leak rate, and spectral radius are $N_x = 400$, $\alpha = 1$, and $\rho (\mathbf{W}) = 0.8$, respectively. 
The grid search for
input scaling ($\delta_{in}$), feedback scaling ($\delta_{fb}$), and the Tikhonov regularization factor $\gamma$ 
was carried out 
as described in Section~\ref{sec:vdp:settings}, but
with a training set ($N_{te}$) of 500 time steps and a validation set ($N_{ve}$) of 300 time steps. 
With the resulting optimal hyperparameter values, $\delta_{fb} = 0.2$, $\delta_{in} = 0.1$, $\gamma = 10^{-5}$ found in the grid search, another ESN was then trained using the total training set of $N_{t} = N_{te} + N_{ve} = 800$ time steps.
Subsequently, the resulting ESN's weights were used as initial values for PI-ESN-a. 

\subsubsection{\eric{PI-ESN-a} training}
\label{sec:4tanks:train}
\begin{figure}[h!]
    \centering
    \includegraphics[width = 0.95\textwidth]{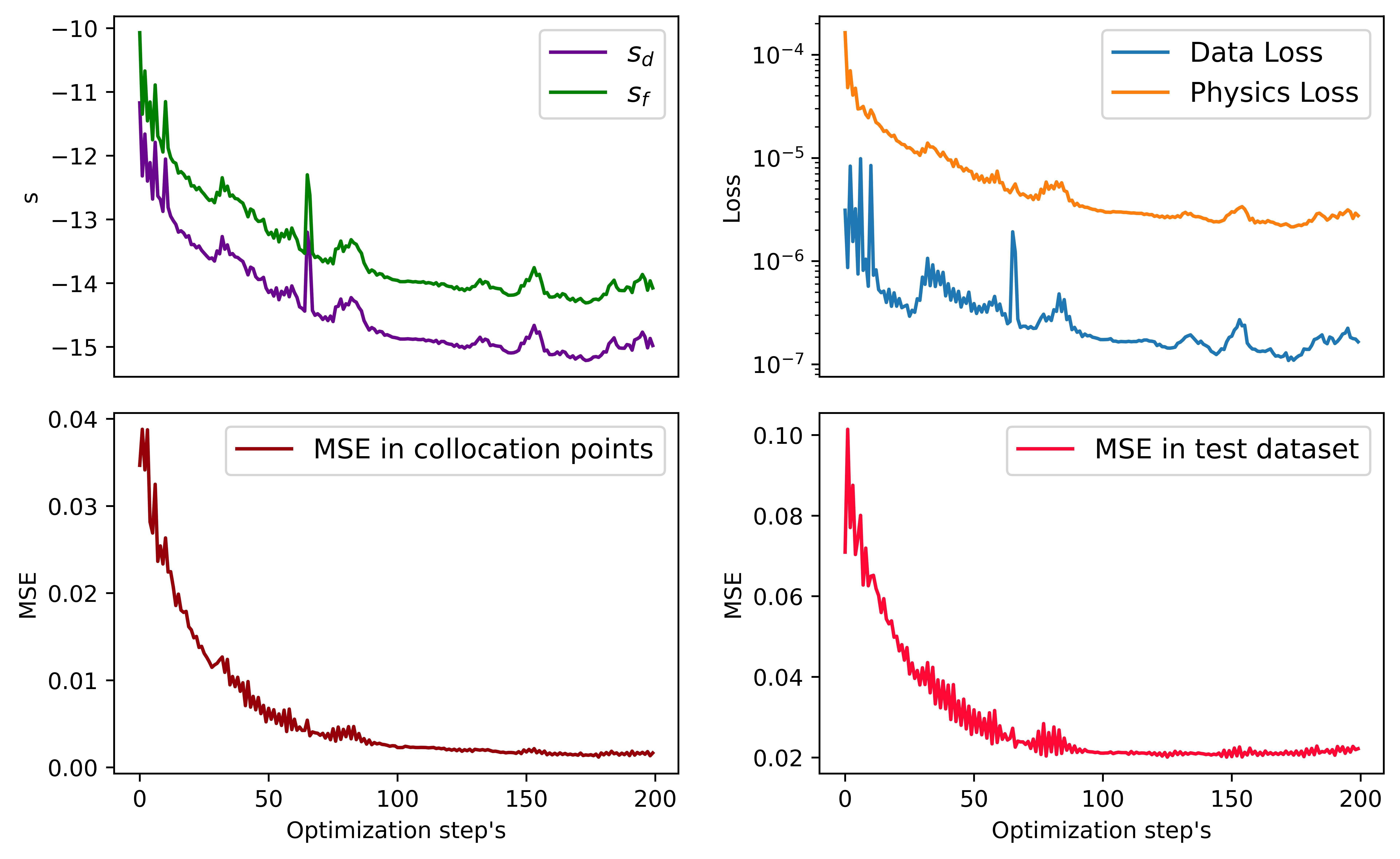}
    \caption{Evolution of the adaptive weights ($s_d, s_f$), the losses functions ($J_{data}, J_{phy}$) and the MSE during the physics training of the PI-ESN-a. 
    } \label{fig:adaptive_4t}
\end{figure}

\begin{figure}[h!]
    \centering
    \includegraphics[width = \textwidth]{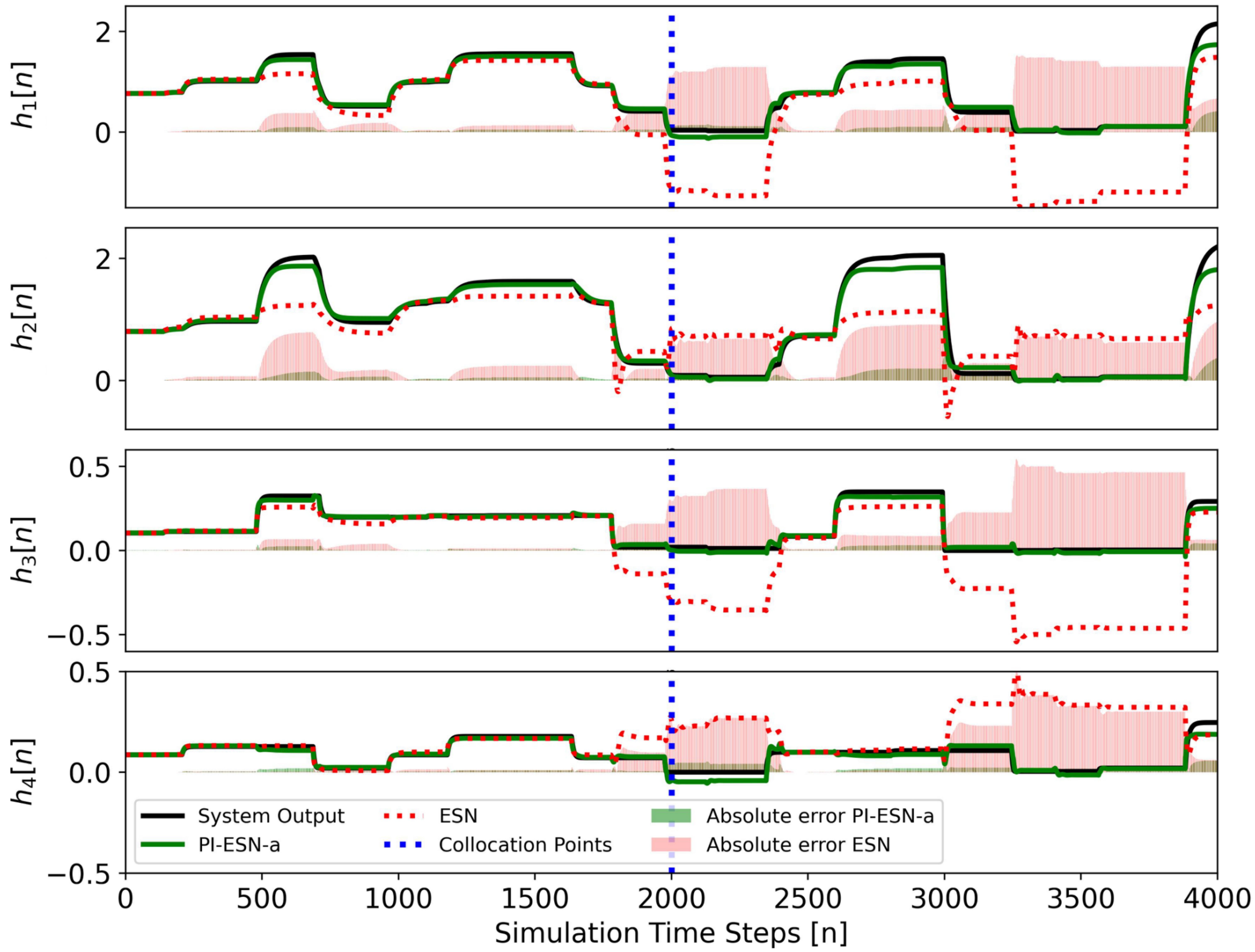}
    \caption{Prediction of the adaptive PI-ESN-a for the four-tank system after training. The blue dashed vertical line splits the region between collocation points (left) and test set (right). In the background, it is possible to observe the absolute error of the ESN and PI-ESN-a with the actual system output. This prediction refers to the collocation points and test data of the system presented in \autoref{fig:fourtanks_sys}. The MSE for the collocation points region was found to be 0.0468 for the ESN and 0.0016 for the PI-ESN-a. In the test region, the corresponding MSE values were 0.1266 and 0.0221, respectively.}
    \label{fig:h1fourtanks_sys}
\end{figure}

The evolution of the PI-ESN-a training process is presented in \autoref{fig:adaptive_4t} in terms of 
the values of adaptive parameters $s_f$ and $s_d$,
of the data loss $J_{data}$ and physics-informed loss $J_{phy}$, regarding the MSE on the collocation points and on the test dataset. 
At the end of the training, 
$J_{data}$ and $J_{phy}$ values were found to be $1.65 \times 10^{-7}$ and $2.76 \times 10^{-6}$, respectively, and $[s_d,s_f]$ parameter values were $[-14.97, -14.07]$. 
The prediction of the PI-ESN-a (in green) and the corresponding ESN (in red) for tank levels $h_1$, $h_2$, $h_3$, and $h_4$ 
is presented in \autoref{fig:h1fourtanks_sys} for the collocation points as well as for the test set.

It can be seen that the PI-ESN-a prediction performance is improved significantly and consistently over the initial ESN, not only on the collocation points (left area of the dashed vertical blue line) but also on new points in the test set generated by random inputs. Thus, for small data regimes, physics-informed training of the ESN is relevant and useful if physical laws are available.


\subsubsection{Effect of labeled dataset size}
\begin{figure}[h]
    \centering
    \includegraphics[width = 0.8\textwidth]{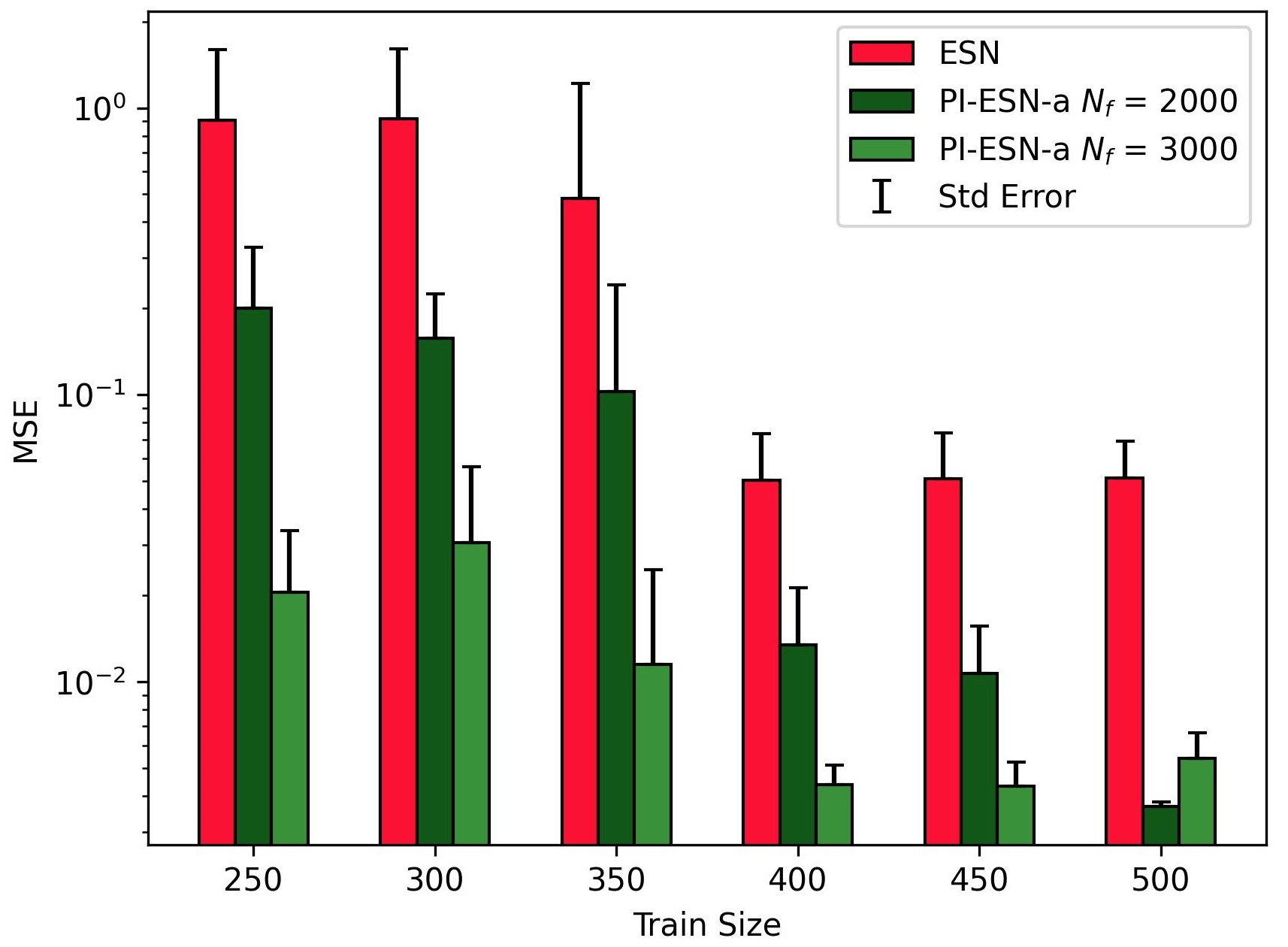}
    \caption{MSE of the PI-ESN-a as a function of the size $N_t$ of the labeled training data set, for two cases $N_f=2000$ and $N_f=3000$, denoting the number of collocation points. The results also include the MSE of a traditional trained ESN for comparison.  Standard deviation is shown for 5 runs with different random seeds, which affect the ESN's initial weights but not the control signal, which remains fixed.}
    \label{fig:PI_experiment_4t}
\end{figure}

To analyze the influence of different training set sizes $N_t$ in a small data regime, an experiment was conducted with PI-ESN-a and its corresponding ESN (Figure \ref{fig:PI_experiment_4t}). 
The prediction MSE over 4000 time steps (collocation points and test set) is averaged over six randomly generated ESNs for each of the six training dataset sizes (250, 300, 350, 400, 450, and 500). 
In addition, the PI-ESN-a experiment was also run for two different numbers of collocation points ($N_f\in \{2000, 3000\}$).

PI-ESN-a was able to reduce the MSE of the original ESN for all training set sizes.
With more data, ESN improves its prediction error, but so does  PI-ESN-a, which shows the potential of the proposed physics-informed ESN training in small data regimes.

\subsubsection{Impact of reservoir Size}
We have also investigated the impact of reservoir size for both PI-ESN-a and corresponding ESN.
Each entry in \autoref{tab:Reservoirexperiment_4t} shows the average MSE over five randomly generated ESNs and four randomly generated inputs for three different reservoir sizes (200, 400, and 800).
In this experiment, a total of 60 runs were conducted. Among them, 6 runs exhibited training instability and were subsequently excluded from the final analysis in the table.

\begin{table}[h!]
\centering
\begin{tabular}{ccccc}
\hline
\multirow{2}{*}{\makecell{Reservoir \\ Size}} & \multicolumn{2}{c}{Collocation Points} & \multicolumn{2}{c}{Test set}   \\ \cline{2-5} 
                                & ESN            & \makecell{PI-ESN-a}       & ESN      & \makecell{PI-ESN-a}  \\ \hline
200                             & 0.415        & 0.032 (-92\%)          & 0.504  & 0.057 (-87\%)     \\
400                             & 0.289        & \textbf{0.028} (-90\%)           & 0.510 & \textbf{0.044}  (-91\%)     \\
800                             & 0.366       & 0.101  (-72\%)          & 0.465 & 0.094 (-80\%)       \\ \hline
\end{tabular}
\caption{\label{tab:Reservoirexperiment_4t} Average MSE for the conventional ESN and PI-ESN-a as a function of the reservoir size ($N_x$) for the four tank system.  Each value is obtained by averaging the MSE of around $20$ experiments.
Between parenthesis, the error reduction percentage of PI-ESN-a over the respective ESN is shown.
}
\end{table}

\autoref{tab:Reservoirexperiment_4t} indicates that irrespective of the reservoir sizes that were investigated, PI-ESN-a reduces the MSE of the original ESN by at least 70\%. The error reduction is bigger, around 90\%, for reservoirs up to 400 neurons.
Additionally, most of the MSE values across various reservoir sizes were on a similar scale, aligning with the behavior observed in the Van der Pol oscillator experiment. 
This behavior can be attributed to the inherent error introduced by the explicit Euler approximation utilized during the physics training process. 
%

\subsubsection{Model Predictive Control}
\color{blue}
{To illustrate how PI-ESN-a can be used in control applications, we present a control experiment in which the levels of tanks 1 and 2 ($h_1$ and $h_2$) are controlled by manipulating the two pump actuators ($V_1$ and $V_2$) of the four-tank plant. 
This application employs Model Predictive Control (MPC), which controls a system using a predictive model to solve an optimization problem in a receding horizon approach at every iteration \cite{Camacho:2007}. 
This section reports results on the four-tank system using the ESN-PNMPC approach proposed by \cite{Jordanou2018-IFAC,Jordanou2021}, which stands for ESNs for Practical Nonlinear Model Predictive Control of dynamical systems.
It is an efficient method that uses a fully nonlinear model to calculate the system's free response and applies a first-order Taylor expansion to approximate the forced response, representing the sensitivity of the response to the control action \cite{Jordanou2021}.
Here, the PI-ESN-a serves as the function mapping the current state (tank levels) and control input (two voltages) to the state at the next time step, ensuring that the MPC predictions satisfy the system dynamics.
} For details on the ESN-PNMPC controller, including system constraints and parameters, see \ref{appendix:ESN-PNMPC}.

\begin{figure}[tb!]
    \centering
    \includegraphics[width = 0.82 \textwidth]{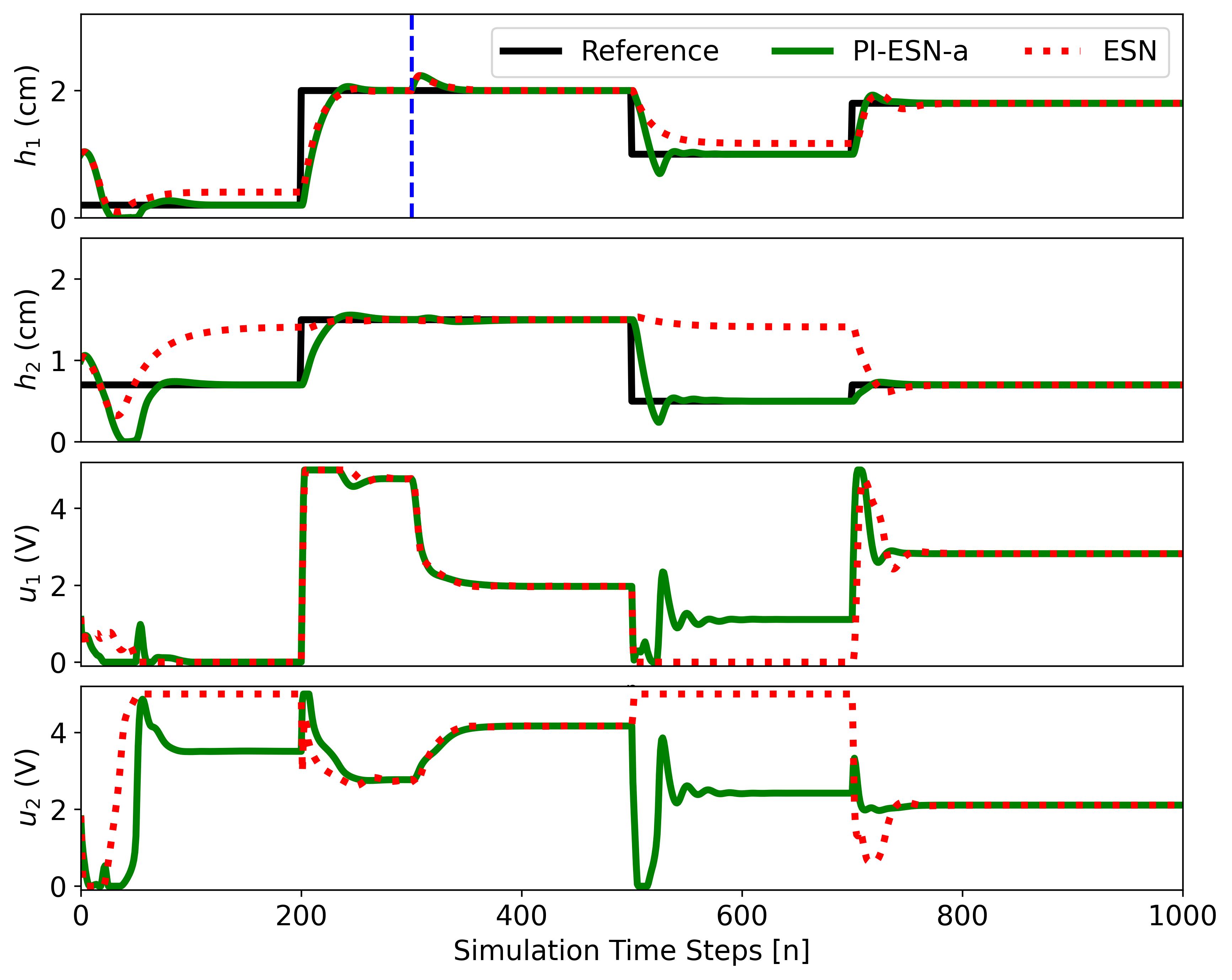}
    \caption{
    \eric{
  Control of the four-tank system via ESN-PNMPC, with two predictive models: PI-ESN-a (in green color) and regular ESN (in red color). 
     The controlled variables are the tank levels $h_1$ and $h_2$, whereas
    the reference trajectory for $h_1$ and $h_2$ corresponds to the black step signal. 
    The control inputs $\mathbf{u}$ are the manipulated voltages shown in the lower plots, found by ESN-PNMPC.
    A perturbation is added at timestep 300, represented by the blue vertical dashed line.
    The tracking error, measured by the IAE between the reference and the respective plant measurement, is 122.68 (423.39) for the control using the PI-ESN-a (regular ESN).
    }
    }
    \label{fig:mpc_fourtanks}
\end{figure}

The PI-ESN-a trained with 500 data samples and 2,000 collocation points was employed as a predictive model in ESN-PNMPC of the four-tank system (Fig. \ref{fig:mpc_fourtanks}). 
In addition, a regular ESN was trained with the same data samples, serving as baseline in the comparison.
Fig. \ref{fig:mpc_fourtanks} shows that the PI-ESN-a works very well as a predictive model in the control of the tank levels, particularly considering a small sample of labeled data (500 timesteps), while previous work utilized 40,000 timesteps for ESN training \cite{Jordanou2021}.

Thus, our ESN with physics-informed training was able to regularize the model, requiring much less training data, while performing well in the control task and, in particular, better than the vanilla ESN-PNMPC in terms of the tracking error: the PNMPC with PI-ESN-a shows an IAE (Integral of Absolute Error) metric (122.68) about 3.45 times lower than the PNMPC with plain ESN (423.39). 
The improved performance is due to the additional training with collocation points and physics laws,
allowing the ESN to generalize to operating points not covered by the training data. 

\subsection{Electric Submersible Pump}
\color{blue}

\subsubsection{Model}
\begin{figure}[htb!]
  \centering
    \includegraphics[width=0.45\linewidth]{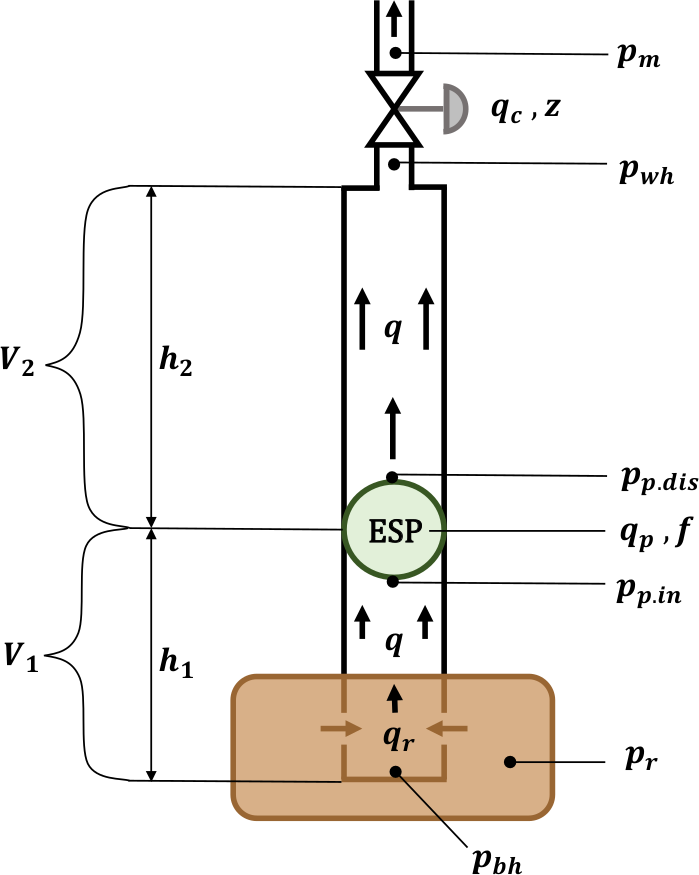}  
    \caption{
    \color{blue}{Schematic of an electric submersible pump. The pressure gradient imposed by the difference between the reservoir pressure $p_r$ and the well bottomhole pressure $p_{bh}$, $(p_r - p_{bh})$, induces the inflow $q_r$ of fluids from the reservoir into the well. The ESP lifts the fluids to the topside by adding energy regulated by the pump frequency $f$, which generates a pressure gain. The wellhead pressure $p_{wh}$ corresponds to the pressure upstream of the production choke, regulated by the choke opening $z$ to ensure pressure balance with the fixed manifold pressure $p_m$.}}
        \label{fig:esp}
\end{figure}

An Electric Submersible Pump (ESP) is used in oil wells to enhance or maintain production, particularly when the reservoir pressure is insufficient to lift fluids to the surface. Fig. \ref{fig:esp} shows the ESP's schematic. 
ESPs are favored for their ability to handle high fluid volumes efficiently with minimal maintenance. They are versatile and suitable for various environments, including onshore and offshore settings, as well as deviated wells. 

The ESP model used here builds on Statoil's (now Equinor) model presented in \cite{PavlovESPmodel}, with viscosity modeling equations from \cite{mheForESP}. It includes an ESP and a production choke valve. 
The basic operation involves the pressure difference between the reservoir pressure $p_r$ and the bottomhole pressure $p_{bh}$, which drives the inflow $q_r$ of fluids into the well. The ESP increases this pressure gradient by adjusting the pump frequency $f$, lifting fluids to the surface. The wellhead pressure $p_{wh}$, regulated by the choke opening $z$, ensures balance with the manifold pressure $p_m$. Operators control the ESP frequency and choke opening to achieve production targets, guided by a reference bottomhole pressure.

The ESP dynamic model includes reservoir inflow, production pipe, ESP, and production choke. Although it omits gas production and viscosity variations, it still accurately represents well dynamics \cite{PavlovESPmodel}. The model has three states: bottomhole pressure $p_{bh}$, wellhead pressure $p_{wh}$, and average flow rate $q$. 
The differential equations are:
\begin{subequations}\label{DiffEqESP}
    \begin{align}
    \dot{p}_{bh} &= \frac{\beta_{1}}{V_1}(q_{r}-q) \\
    \dot{p}_{wh} &= \frac{\beta_{2}}{V_2}(q-q_c) \\
    \dot{q} &= \frac{1}{M}(p_{bh}-p_{wh}-\rho g h_w- \Delta p_f +\Delta P_p)
    \end{align}
\end{subequations}
\eric{where $\Delta p_f$ represents the frictional pressure loss, $\Delta P_p$ is the pressure gain from the ESP, $h_w$ is the well's vertical length, $\rho$ is the fluid density, and $g$ is gravitational acceleration. These equations, along with algebraic constraints, form a Differential Algebraic Equation (DAE) system. The relationship between pump frequency $f$ and ESP pressure gain $\Delta P_p$ is highly nonlinear, as is the flow $q_c$ through the choke, which depends on bottomhole pressure, manifold pressure, and choke opening.}

\eric{For a detailed description of the ESP model's variables, parameters, and algebraic equations, refer to \ref{appendix:ESP}.}



\subsubsection{Dataset}

\begin{figure}[tb!]
    \centering
    \includegraphics[width = 0.9\textwidth]{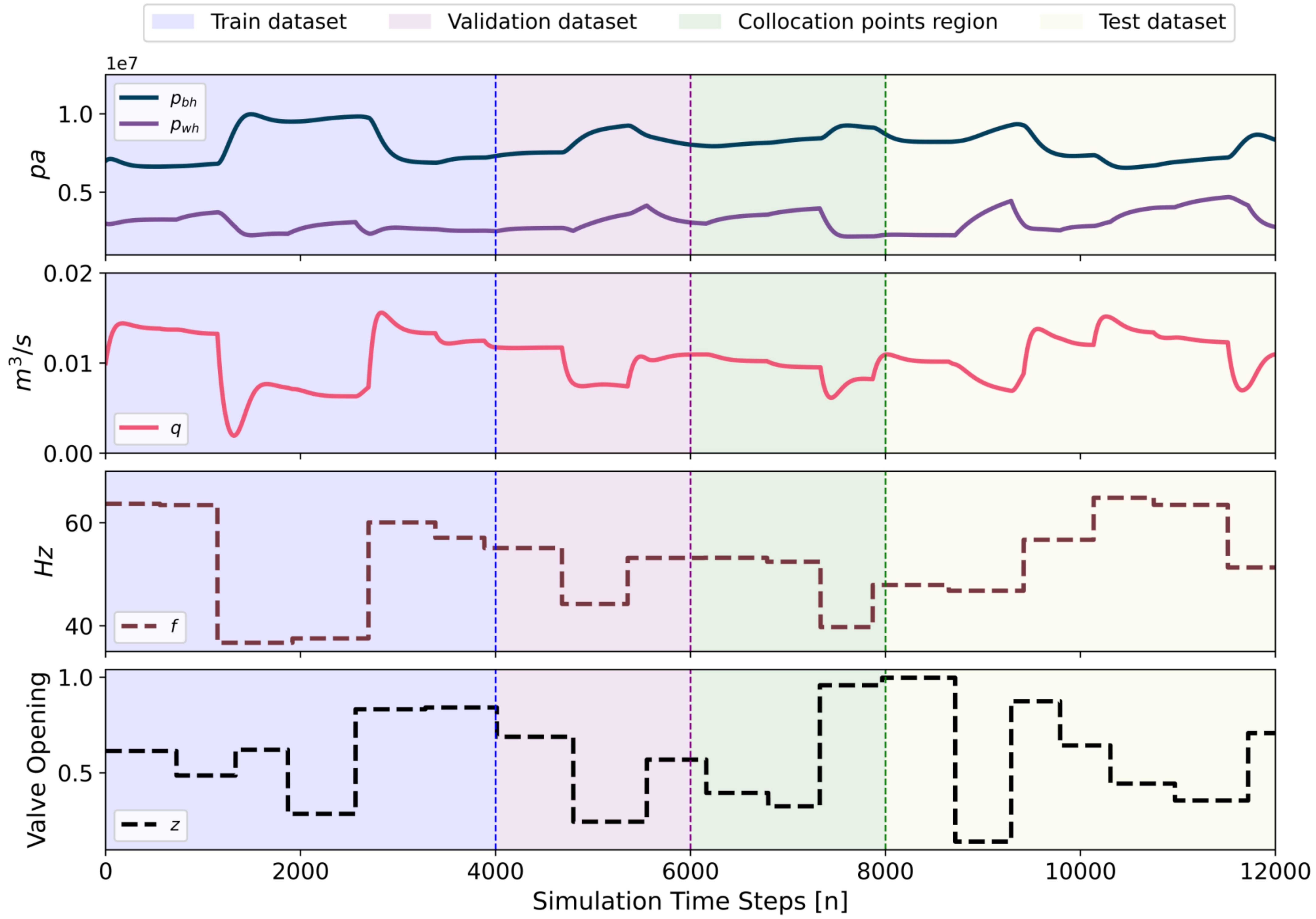}
     \caption{
     \color{blue}
     Dataset for the Electric Submersible Pump. The two upper plots present the three state variables, while the two lower plots show the randomly generated control inputs.
     The training set, consisting of 4,000 points, and the validation set, with 2,000 points, are divided by a dashed blue vertical line. A dashed purple vertical line separates the labeled data from the 2,000 unlabeled collocation points. Finally, the test set, containing 4,000 points, is used for evaluation.
     }
     \label{fig:ESP_simulated_system}
\end{figure}

To simulate the system, the Gekko library \cite{Beal2018gekko} was employed, which is a Python package primarily designed for solving dynamic optimization and control problems.
%
The ESP system was simulated using a time step of $\Delta t = 0.01 \ s$. The initial conditions for the system were $p_{bh} = 70 \times 10^6 \ pa$, $p_{wh} = 20 \times 10^6 \ pa$, and $q = 0.01 \ m^3/s$. The dataset generated by the simulation is shown in Fig. \ref{fig:ESP_simulated_system}.

An APRBS input signal was generated with values for
the choke valve opening $z$ 
ranging from 0.1 to 1 and 
the ESP frequency $f$ 
ranging from 35 to 65 Hz, considering a signal variation occurring every 500 to 800 time steps. 
In Fig. \ref{fig:ESP_simulated_system}, the training (validation) set consists of $N_{te} = 4,000$ ($N_{ve} = 2,000$) time steps. 
These two sets are separated by a vertical dashed blue line.
The validation set is used for hyperparameter tuning during the ESN training. 
A vertical dashed purple line splits the dataset between training data ($N_t = 6,000$ time steps) and (unlabeled) collocation points ($N_f = 2,000$ time steps).
Finally, the test data comprises $4,000$ time steps. The data were normalized for ESN training using min-max scaling.

\subsubsection{PI-ESN-a settings and results}

\begin{figure}[tb!]
    \centering
    \includegraphics[width = 0.8\textwidth]{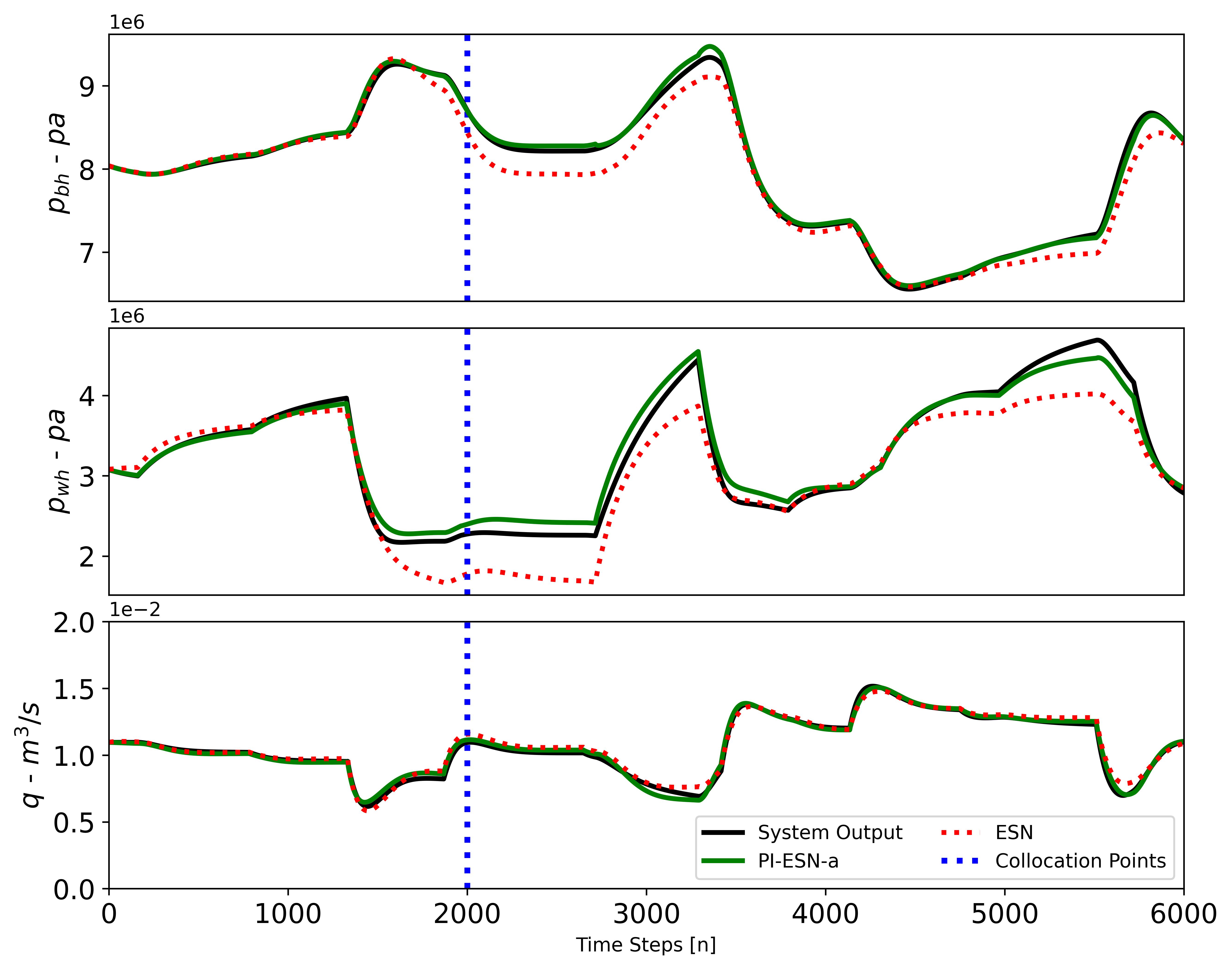}
    \caption{
    \color{blue}
    Prediction of ESN (in red) and PI-ESN-a (in green) for the ESP system after training.
    In solid black line, the (unknown) target values are shown. 
    From 0 to $2,000$ steps, the predictions for the collocation points are presented. After $2,000$ steps, the predictions for the test set are showcased.}
    \label{fig:h1ESP}
\end{figure}

The ESN was initially configured with hyperparameters close to those suggested in \cite{JORDANOU2022}, utilizing $N_x = 300$ and a warm-up period of 50 time steps. Bayesian optimization was employed to fine-tune the hyperparameters $\alpha$, $\rho (\mathbf{W})$, $\delta_{b}$, $\delta_{fb}$, and $\delta_{in}$ using the \textit{BayesianOptimization} package in Python \cite{Stander2002}. This choice of optimization method was motivated by the number of hyperparameters involved.
Subsequently, the ESN was retrained with a total of $N_{t} = 6000$ time steps, utilizing the following values: $\delta_{fb} = 0.1$, $\delta_{in} = 0.1$, $\delta_{b} = 0.1$, $\gamma = 0.0599$, $\alpha = 0.15$, and $\rho (\mathbf{W}) = 0.8$, as determined through the Bayesian search.  
After physics-informed training using data points and collocation points, we evaluate the PI-ESN-a predictions for the collocation points as well as for the test set, as depicted in Fig. \ref{fig:h1ESP}, where
a vertical dashed blue line splits both regions.
The MSE for the collocation points (test set) region was found to be 0.0026 (0.0084) for the ESN and 0.0004 (0.0011) for the PI-ESN-a (Table~\ref{tab:comparison_mse_esp}), showing a reduction of 86.9\% in the MSE of the test set when using the PI-ESN-a.

A comparison of execution times for computing predictions on the test dataset was conducted between the dynamic simulator Gekko and the trained PI-ESN-a model, using 10 runs. On average, PI-ESN-a completed the task in 3.9 seconds, while the Gekko optimizer required 98.5 seconds. The experiment was performed on an Intel(R) Core(TM) i7-9750H CPU @ 2.60GHz, with a clock speed of 2592 MHz, 6 cores, and 12 logical processors. These results demonstrate that PI-ESN-a is approximately 96.32\% faster than the numerical solution obtained using Gekko for solving the DAE.


\begin{table}[tbp]
    \centering
    \color{blue}
    \begin{tabular}{|l|c|c|c|}    
        \hline
        \textbf{Metric} & \textbf{ESN} & \textbf{PI-ESN-a} & \textbf{Reduction} \\
        \hline
        MSE (Collocation Points) & 0.0026 & 0.0004 & 84.62\% \\
        \hline
        MSE (Test Dataset) & 0.0084 & 0.0011 & 86.90\% \\
        \hline
    \end{tabular}
    \caption{Test MSE for ESN and PI-ESN-a on the Electric Submersible Pump model.}    
       \label{tab:comparison_mse_esp}
\end{table}

\color{black}

\section{Conclusion}
\label{sec:conclusion}
In this work, we have proposed an extension of Physics-Informed Echo State Networks (PI-ESN) that make them work with external inputs. 
This augmentation allows PI-ESNs to be controlled through manipulation of their input, which is useful, for example, in Model Predictive Control applications of industrial plants. 
Additionally, 
we have enhanced the PI-ESN with external inputs by incorporating a 
self-adaptive balancing loss method, initially developed for PINNs. 
The resulting framework, PI-ESN-a, has enabled dynamic balancing of the significance of both the data and physics-informed loss terms within the total loss function, achieved through adaptation of scaling parameters during training.

Our PI-ESN-a was shown to perform better than the corresponding ESN (with weights equivalent to the pretrained PI-ESN-a) in all of the investigated applications, the Van der Pol oscilator, the four-tank system and the \eric{electric submersible pump}, with relative error reduction up to 92\%. 
This is particularly valid for small data regimes, where the number of labeled samples is limited and a priori information on the differential equations of the system is available. 
\eric{In particular, MPC using the PI-ESN-a for the four-tank system was shown to perform significantly better than MPC using the plain ESN, showing a relative error reduction of about 71\%. 
}

Upcoming work will tackle the modeling with PI-ESN-a and control of more complex, challenging systems, such as industrial plants.
Besides, we will investigate other ways to compute the loss function gradient to improve the training stability in special scenarios in which the predicted output feedback into the reservoir may severely affect the gradient computation. Extensions of the approach applicable to Partial Differential Equations is another important topic for future research.


\section*{Acknowledgements}

The authors would like to express their gratitude to the Human Resources Program of the National Agency of Petroleum, Natural Gas, and Biofuels (PRH-ANP) and FAPESC (grant 2021TR2265) for financial support.

\bibliographystyle{elsarticle-num} 
\bibliography{mybib}

\begin{thebibliography}{10}
\expandafter\ifx\csname url\endcsname\relax
  \def\url#1{\texttt{#1}}\fi
\expandafter\ifx\csname urlprefix\endcsname\relax\def\urlprefix{URL }\fi
\expandafter\ifx\csname href\endcsname\relax
  \def\href#1#2{#2} \def\path#1{#1}\fi

\bibitem{Raissi2019}
M.~Raissi, P.~Perdikaris, G.~E. Karniadakis, Physics-informed neural networks: A deep learning framework for solving forward and inverse problems involving nonlinear partial differential equations, Journal of Computational Physics 378 (2019) 686--707.
\newblock \href {https://doi.org/10.1016/j.jcp.2018.10.045} {\path{doi:10.1016/j.jcp.2018.10.045}}.

\bibitem{Karniadakis2021}
G.~E. Karniadakis, I.~G. Kevrekidis, L.~Lu, P.~Perdikaris, S.~Wang, L.~Yang, Physics-informed machine learning, Nature Reviews Physics 3~(6) (2021) 422--440.
\newblock \href {https://doi.org/10.1038/s42254-021-00314-5} {\path{doi:10.1038/s42254-021-00314-5}}.

\bibitem{Edwards2022}
C.~Edwards, Neural networks learn to speed up simulations, Communications of ACM 65~(5) (2022) 27–29.
\newblock \href {https://doi.org/10.1145/3524015} {\path{doi:10.1145/3524015}}.

\bibitem{Eric2021}
E.~A. Antonelo, E.~Camponogara, L.~O. Seman, E.~R. de~Souza, J.~P. Jordanou, J.~F. Hubner, Physics-informed neural nets for control of dynamical systems, Neurocomputing 579 (2024).
\newblock \href {https://doi.org/10.1016/j.neucom.2024.127419} {\path{doi:10.1016/j.neucom.2024.127419}}.

\bibitem{Jaeger2001a}
H.~Jaeger, \href{https://www.ai.rug.nl/minds/uploads/EchoStatesTechRep.pdf}{The {\textquoteleft}{\textquoteleft}echo state{\textquoteright}{\textquoteright} approach to analysing and training recurrent neural networks -- with an erratum note}, Tech. Rep. GMD 148, Fraunhofer Institute for Autonomous Intelligent Systems (2001).
\newline\urlprefix\url{https://www.ai.rug.nl/minds/uploads/EchoStatesTechRep.pdf}

\bibitem{Verstraeten2006a}
D.~Verstraeten, B.~Schrauwen, M.~D'Haene, D.~Stroobandt, An experimental unification of reservoir computing methods, Neural Networks 20~(3) (2007) 391--403.
\newblock \href {https://doi.org/10.1016/j.neunet.2007.04.003} {\path{doi:10.1016/j.neunet.2007.04.003}}.

\bibitem{Antonelo2014}
E.~Antonelo, B.~Schrauwen, On learning navigation behaviors for small mobile robots with reservoir computing architectures, IEEE Transactions on Neural Networks and Learning Systems 26~(4) (2014) 763--780.
\newblock \href {https://doi.org/10.1109/TNNLS.2014.2323247} {\path{doi:10.1109/TNNLS.2014.2323247}}.

\bibitem{Zhou2022}
J.~Zhou, T.~Han, F.~Xiao, G.~Gui, B.~Adebisi, H.~Gacanin, H.~Sari, Multiscale network traffic prediction method based on deep echo-state network for internet of things, IEEE Internet of Things Journal 9~(21) (2022) 21862--21874.
\newblock \href {https://doi.org/10.1109/JIOT.2022.3181807} {\path{doi:10.1109/JIOT.2022.3181807}}.

\bibitem{ROBERTS2022}
C.~Roberts, J.~D. Lara, R.~Henriquez-Auba, M.~Bossart, R.~Anantharaman, C.~Rackauckas, B.-M. Hodge, D.~S. Callaway, Continuous-time echo state networks for predicting power system dynamics, Electric Power Systems Research 212 (2022) 108562.
\newblock \href {https://doi.org/10.1016/j.epsr.2022.108562} {\path{doi:10.1016/j.epsr.2022.108562}}.

\bibitem{Jordanou2019}
J.~P. Jordanou, E.~A. Antonelo, E.~Camponogara, Online learning control with echo state networks of an oil production platform, Engineering Applications of Artificial Intelligence 85 (2019) 214--228.
\newblock \href {https://doi.org/10.1016/j.engappai.2019.06.011} {\path{doi:10.1016/j.engappai.2019.06.011}}.

\bibitem{SHAHI2022}
S.~Shahi, F.~H. Fenton, E.~M. Cherry, Prediction of chaotic time series using recurrent neural networks and reservoir computing techniques: A comparative study, Machine Learning with Applications 8 (2022) 100300.
\newblock \href {https://doi.org/10.1016/j.mlwa.2022.100300} {\path{doi:10.1016/j.mlwa.2022.100300}}.

\bibitem{Ceni2024residual}
A.~Ceni, C.~Gallicchio, Residual echo state networks: Residual recurrent neural networks with stable dynamics and fast learning, Neurocomputing (2024) 127966.

\bibitem{Micheli2022discrete}
A.~Micheli, D.~Tortorella, Discrete-time dynamic graph echo state networks, Neurocomputing 496 (2022) 85--95.

\bibitem{Doan2019}
N.~A.~K. Doan, W.~Polifke, L.~Magri, Physics-informed echo state networks for chaotic systems forecasting, in: International Conference on Computational Science, Springer, 2019, pp. 192--198.
\newblock \href {https://doi.org/10.1007/978-3-030-22747-0_15} {\path{doi:10.1007/978-3-030-22747-0_15}}.

\bibitem{Camacho:2007}
E.~F. Camacho, C.~Bordon, Model Predictive Control, Springer, 2007.

\bibitem{Xiang2022}
Z.~Xiang, W.~Peng, X.~Liu, W.~Yao, Self-adaptive loss balanced physics-informed neural networks, Neurocomputing 496 (2022) 11--34.
\newblock \href {https://doi.org/10.1016/j.neucom.2022.05.015} {\path{doi:10.1016/j.neucom.2022.05.015}}.

\bibitem{Jordanou2021}
J.~P. Jordanou, E.~A. Antonelo, E.~Camponogara, Echo state networks for practical nonlinear model predictive control of unknown dynamic systems, IEEE Transactions on Neural Networks and Learning Systems 33~(6) (2021) 2615--2629.
\newblock \href {https://doi.org/10.1109/TNNLS.2021.3136357} {\path{doi:10.1109/TNNLS.2021.3136357}}.

\bibitem{Pathak_2018}
J.~Pathak, A.~Wikner, R.~Fussell, S.~Chandra, B.~R. Hunt, M.~Girvan, E.~Ott, Hybrid forecasting of chaotic processes: Using machine learning in conjunction with a knowledge-based model, Chaos: An Interdisciplinary Journal of Nonlinear Science 28~(4) (2018) 041101.
\newblock \href {https://doi.org/10.1063/1.5028373} {\path{doi:10.1063/1.5028373}}.

\bibitem{doan20192}
N.~A.~K. Doan, W.~Polifke, L.~Magri, A physics-aware machine to predict extreme events in turbulence (2019).
\newblock \href {http://arxiv.org/abs/1912.10994} {\path{arXiv:1912.10994}}.

\bibitem{Doan2020}
N.~A.~K. Doan, W.~Polifke, L.~Magri, Learning hidden states in a chaotic system: A physics-informed echo state network approach (2020).
\newblock \href {https://doi.org/10.48550/arXiv.2101.00002} {\path{doi:10.48550/arXiv.2101.00002}}.

\bibitem{Racca2021}
A.~Racca, L.~Magri, Automatic-differentiated physics-informed echo state network {(API-ESN)}, arXiv preprint arXiv:2101.00002 (2021).
\newblock \href {https://doi.org/10.48550/arXiv.2101.00002} {\path{doi:10.48550/arXiv.2101.00002}}.

\bibitem{oh2023pure}
D.~K. Oh, Pure physics-informed echo state network of {ODE} solution replicator, in: International Conference on Artificial Neural Networks, Springer, 2023, pp. 225--236.
\newblock \href {https://doi.org/10.1007/978-3-031-44201-8_19} {\path{doi:10.1007/978-3-031-44201-8_19}}.

\bibitem{Yildiz2012}
I.~B. Yildiz, H.~Jaeger, S.~J. Kiebel, Re-visiting the echo state property, Neural Networks 35 (2012) 1--9.
\newblock \href {https://doi.org/10.1016/j.neunet.2012.07.005} {\path{doi:10.1016/j.neunet.2012.07.005}}.

\bibitem{Lukoševičius2012}
M.~Luko{\v{s}}evi{\v{c}}ius, A Practical Guide to Applying Echo State Networks, Springer Berlin Heidelberg, Berlin, Heidelberg, 2012, pp. 659--686.

\bibitem{Kingma2014}
D.~P. Kingma, J.~Ba, {ADAM}: A method for stochastic optimization, arXiv preprint arXiv:1412.6980 (2014).
\newblock \href {https://doi.org/10.48550/arXiv.1412.6980} {\path{doi:10.48550/arXiv.1412.6980}}.

\bibitem{Andrew2007}
G.~Andrew, J.~Gao, Scalable training of l1-regularized log-linear models, in: Proceedings of the 24th International Conference on Machine Learning, ICML '07, Association for Computing Machinery, New York, NY, USA, 2007, p. 33–40.
\newblock \href {https://doi.org/10.1145/1273496.1273501} {\path{doi:10.1145/1273496.1273501}}.

\bibitem{Bishop2006}
C.~M. Bishop, Pattern Recognition and Machine Learning (Information Science and Statistics), Springer-Verlag Inc., New York, 2006.

\bibitem{Hafeez2015vdp}
H.~Y. Hafeez, C.~E. Ndikilar, S.~Isyaku, Analytical study of the van der pol equation in the autonomous regime, Progress in Physics 11 (2015) 252--255.

\bibitem{Tsatsos2008}
M.~Tsatsos, The van der pol equation, arXiv preprint arXiv:0803.1658 (2008).
\newblock \href {https://doi.org/10.48550/ArXiv.0803.1658} {\path{doi:10.48550/ArXiv.0803.1658}}.

\bibitem{ALVARADO2006}
I.~Alvarado, D.~Limon, W.~García-Gabín, T.~Alamo, E.~Camacho, An educational plant based on the quadruple-tank process, IFAC Proceedings Volumes 39~(6) (2006) 82--87, 7th IFAC Symposium on Advances in Control Education.
\newblock \href {https://doi.org/10.3182/20060621-3-ES-2905.00016} {\path{doi:10.3182/20060621-3-ES-2905.00016}}.

\bibitem{Johansson2000}
K.~H. Johansson, The quadruple-tank process: A multivariable laboratory process with an adjustable zero, IEEE Transactions on Control Systems Technology 8~(3) (2000) 456--465.
\newblock \href {https://doi.org/10.3182/20060621-3-ES-2905.00016} {\path{doi:10.3182/20060621-3-ES-2905.00016}}.

\bibitem{Jordanou2018-IFAC}
J.~P. Jordanou, E.~Camponogara, E.~A. Antonelo, M.~A.~S. Aguiar, Nonlinear model predictive control of an oil well with echo state networks, IFAC Proceedings Volumes 51~(8) (2018) 13--18.
\newblock \href {https://doi.org/10.1016/j.ifacol.2018.06.348} {\path{doi:10.1016/j.ifacol.2018.06.348}}.

\bibitem{PavlovESPmodel}
A.~{Pavlov}, D.~{Krishnamoorthy}, K.~{Fjalestad}, E.~{Aske}, M.~{Fredriksen}, Modelling and model predictive control of oil wells with electric submersible pumps, in: IEEE Conference on Control Applications (CCA), 2014, pp. 586--592.
\newblock \href {https://doi.org/10.1109/CCA.2014.6981403} {\path{doi:10.1109/CCA.2014.6981403}}.

\bibitem{mheForESP}
B.~J. Binder, A.~Pavlov, T.~A. Johansen, Estimation of flow rate and viscosity in a well with an electric submersible pump using moving horizon estimation, Vol.~48, IFAC-PapersOnLine, 2015, pp. 140--146.
\newblock \href {https://doi.org/10.1016/j.ifacol.2015.08.022} {\path{doi:10.1016/j.ifacol.2015.08.022}}.

\bibitem{Beal2018gekko}
L.~D. Beal, D.~C. Hill, R.~A. Martin, J.~D. Hedengren, {GEKKO} optimization suite, Processes 6~(8) (2018) 106.
\newblock \href {https://doi.org/10.3390/pr6080106} {\path{doi:10.3390/pr6080106}}.

\bibitem{JORDANOU2022}
J.~P. Jordanou, I.~Osnes, S.~B. Hernes, E.~Camponogara, E.~A. Antonelo, L.~Imsland, Nonlinear model predictive control of electrical submersible pumps based on echo state networks, Advanced Engineering Informatics 52 (2022) 101553.
\newblock \href {https://doi.org/10.1016/j.aei.2022.101553} {\path{doi:10.1016/j.aei.2022.101553}}.

\bibitem{Stander2002}
N.~Stander, K.~Craig, On the robustness of a simple domain reduction scheme for simulation-based optimization, International Journal for Computer-Aided Engineering and Software (Eng. Comput.) 19 (06 2002).
\newblock \href {https://doi.org/10.1108/02644400210430190} {\path{doi:10.1108/02644400210430190}}.

\bibitem{PLUCENIO2007210}
A.~Plucenio, D.~Pagano, A.~Bruciapaglia, J.~Normey-Rico, A practical approach to predictive control for nonlinear processes, IFAC Proceedings Volumes 40~(12) (2007) 210--215, 7th IFAC Symposium on Nonlinear Control Systems.
\newblock \href {https://doi.org/10.3182/20070822-3-ZA-2920.00035} {\path{doi:10.3182/20070822-3-ZA-2920.00035}}.

\end{thebibliography}

\appendix

\section{ESN-PNMPC}\label{appendix:ESN-PNMPC}
\color{blue}


The Practical Nonlinear Model Predictive Controller (PNMPC) offers a method for decomposing a nonlinear model into a free response and a forced response, using a first-order Taylor expansion \cite{PLUCENIO2007210}. The ESN-PNMPC utilizes the neural network as a model to compute the predictions. Assuming a dynamic system in the form:
\begin{equation}
\begin{aligned}
\mathbf{x}[k+i] & =\mathbf{f}(\mathbf{x}[k+i-1], \mathbf{u}[k+i-1]),  \\
\mathbf{y}[k+i] & =\mathbf{g}(\mathbf{x}[k+i]),  \\
\mathbf{u}[k+i-1]) & =\mathbf{u}[k-1]+\sum_{j=0}^{i-1} \Delta \mathbf{u}[k+j], 
\end{aligned}
\end{equation}
where \( f(\cdot) \) and \( g(\cdot) \) are given by the Equations \ref{eq:stateupESN} and \ref{eq:outputESN}, respectively. The prediction vector in PNMPC is calculated as follows:
\begin{equation}
   \widehat{\mathbf{Y}} =\mathbf{G} \cdot \boldsymbol{\Delta} \mathbf{U}+\mathbf{F}, \label{PNMPC:App:eq:pred}
\end{equation}
\begin{equation}
\boldsymbol{\Delta} \mathbf{U} =\left(\begin{array}{c}
\boldsymbol{\Delta} \mathbf{u}[k] \\
\boldsymbol{\Delta}[k+1] \\
\vdots \\
\boldsymbol{\Delta} \mathbf{u}\left[k+N_{u}-1\right]
\end{array}\right),
\end{equation}
\begin{equation}
\mathbf{F} =\left(\begin{array}{c}
\mathbf{g}(\mathbf{f}(\mathbf{x}[k], \mathbf{u}[k-1])) \\
\mathbf{g}(\mathbf{f}(\mathbf{x}[k+1], \mathbf{u}[k-1])) \\
\vdots \\
\mathbf{g}\left(\mathbf{f}\left(\mathbf{x}\left[k+N_{y}-1\right], \mathbf{u}[k-1]\right)\right)
\end{array}\right),
\end{equation}
\begin{equation}
\mathbf{G}=\left(\begin{array}{cccc}
\frac{\partial \mathbf{y}[k+1]}{\partial \mathbf{u}[k]} & 0 & \cdots & 0 \\
\frac{\partial \mathbf{y}[k+2]}{\partial \mathbf{u}[k]} & \frac{\partial \mathbf{y}[k+2]}{\partial \mathbf{u}[k+1]} & \cdots & 0 \\
\vdots & \vdots & \ddots & \vdots \\
\frac{\partial \mathbf{y}\left[k+N_{y}\right]}{\partial \mathbf{u}[k]} & \frac{\partial \mathbf{y}\left[k+N_{y}\right]}{\partial \mathbf{u}[k+1]} & \cdots & \frac{\partial \mathbf{y}\left[k+N_{y}\right]}{\partial \mathbf{u}\left[k+N_{u}-1\right]}
\end{array}\right),
\end{equation}
where $N_{y}$ is the prediction horizon and $N_{u}$ is the control horizon. The vector $\boldsymbol{\Delta} \mathbf{U}$ consists of the control increment values concatenated along $N_{u}$. In the PNMPC, a low-pass filter was employed to reject the disturbances and errors between the system's output and the model's prediction. Specifically, the filter was applied to the prediction error vector \( n[k] \). The free response with the low-pass filter can be expressed as:
\begin{align}
\mathbf{F}&=\left[\begin{array}{c}
\mathbf{g}(\mathbf{f}(x[k], u[k-1])) \\
\mathbf{g}(\mathbf{f}(x[k+1], u[k-1])) \\
\vdots \\
\mathbf{g}\left(\mathbf{f}\left(x\left[k+N_y-1\right], u[k-1]\right)\right)
\end{array}\right]+n[k] \\
n[k]&=n[k-1]+ \Delta n[k] \\
\Delta n[k]&=(1-b)\left(\widehat{y}[k \mid k-1]-y_m[k]\right)+b \Delta n[k-1] \\
\widehat{y}[k \mid k-1]&=g(f(x[k-1], u[k-1]))+n[k-1],
\end{align}
where the parameter \( b \in [0, 1) \) sets the cutoff frequency. This parameter was adjusted to optimize the trade-off between disturbance rejection and system robustness.

The system is linearized only with respect to the control input so that the nonlinear term 
$\mathbf{F}$ is derived under the assumption that the previous control action $\mathbf{u}[k-1]$ remains constant. The forced response of the system is then determined by multiplying the sensitivity matrix $\mathbf{G}$, which comprises the system’s Jacobians, by the control increment vector $\mathbf{\Delta U}$ across the prediction horizon. This approach effectively transforms the Model Predictive Control (MPC) problem into a Quadratic Programming (QP) problem, enabling the application of efficient optimization techniques in real-time control scenarios.

The calculation of $\mathbf{F}$ is relatively straightforward, involving direct function evaluation. However, the computation of the sensitivity matrix $\mathbf{G}$ is more complex. Earlier studies, such as \cite{PLUCENIO2007210}, operated under the assumption that Jacobians were unobtainable, resorting to finite-difference schemes. While effective, this approach could result in significant computational complexity, particularly for multivariate systems. In contrast, this work leverages a known state equation model, specifically an Echo State Network (ESN), allowing the controller to employ a recursive strategy for calculating 
$\mathbf{G}$ by applying the chain rule, thereby enhancing computational efficiency. The algorithm for computing the sensitivities is detailed in \cite{JORDANOU2022}.

By applying the prediction model \eqref{PNMPC:App:eq:pred} in an MPC framework, the following quadratic program is solved at each time step: 
\begin{equation}
\begin{aligned}
\underset{\mathbf{\Delta U}}{\min} ~~ & J(\mathbf{\Delta U}) = \mathbf{\Delta U}^T \mathbf{H} \mathbf{\Delta U} + \mathbf{c}^T \mathbf{\Delta U} \\
   \text{s.t.}:~ & \mathbf{T}\mathbf{\Delta U} \leq \mathbf{1}\mathbf{u}^{\max}  - \mathbf{1}\mathbf{u}[k - 1]  \\
    & \mathbf{T}\mathbf{\Delta U} \geq \mathbf{1}\mathbf
{u}^{\min} - \mathbf{1}\mathbf{u}[k - 1] \\
    & \mathbf{G}\mathbf{\Delta U} \leq \mathbf{1}  \otimes \mathbf{y}^{\max} - \mathbf{F}  \\
    & \mathbf{G}\mathbf{\Delta U} \geq \mathbf{1}  \otimes \mathbf{y}^{\min} - \mathbf{F} 
\end{aligned}
\end{equation}
where:
\begin{align*}
   \mathbf{H} &= \mathbf{G}^T \mathbf{Q}\mathbf{G} + \mathbf{R} \\
    \mathbf{c} &= 2\mathbf{G}^{T} \mathbf{Q}^{T} (\mathbf{Y}_{\text{ref}} - \mathbf{F})
\end{align*}
and $\otimes$ is the Kronecker product, $\mathbf{Q}$ is a positive semidefinite matrix penalizing deviation from reference, $\mathbf{R}$ is a positive definite matrix penalizing control action variation, and $\mathbf{Y}_{\text{ref}}$ is the output reference over the prediction horizon, $\mathbf{u}^{\min}$ and $\mathbf{u}^{\max}$ define bounds on control inputs, and $\mathbf{y}^{\min}$ and $\mathbf{y}^{\max}$ impose bounds on system outputs.
\color{black}



\color{blue}
For the four-tank system, the prediction horizon \( N_y \) was set to 10 time steps and the control horizon \( N_u \) was set to 3 time steps. The identity matrix was used for both error weights \( \mathbf{Q} \) and control variation weights \( \mathbf{R} \), with the weights set to 5 and 1, respectively. This means that the two reference errors were penalized 5 times more than the control effort. The filter parameter \( b \) was set to 0.6. Control inputs were constrained with bounds \( \mathbf{u}^{\min} = 0 \) V and \( \mathbf{u}^{\max} = 5 \) V, and system outputs were bounded by \( \mathbf{y}^{\min} = 0 \) cm and \( \mathbf{y}^{\max} = 3 \) cm.

\color{black}


\section{ESP Algebraic Equations}\label{appendix:ESP}

\bb{The algebraic equations associated with the dynamic equations of the ESP model are presented below according to their properties.}
\begin{itemize}
\item \bb{Flow equations:
   \begin{subequations}
    \begin{align}
        q_r &= PI(p_r - p_{bh}) \\
        q_c &= C_c\,z \sqrt{p_{wh}-p_m} 
    \end{align}
\end{subequations}}

\item \bb{Friction equations:
\begin{subequations}
    \begin{align}
        \Delta p_f &= F_1 + F_2 \\
        F_i &= 0.158 \; \frac{\rho L_i q^2}{D_i A_i^2}\left(\frac{\mu}{\rho D_i q} \right) ^{\frac{1}{4}}
    \end{align}
\end{subequations}}

\item \bb{ESP equations:
\begin{subequations}
    \begin{align}
    \Delta p_p &= \rho  g  H \\
    H &= C_H(\mu) \left(c_0 + c_1  \left(\frac{q}{C_Q(\mu)} \ \frac{f_0}{f} \right) - c_2 \left(\frac{q}{C_Q(\mu)} \ \frac{f_0}{f} \right)^2  \left(\frac{f}{f_0} \right)^2 \right) \\
    c_0 &= 9.5970\cdot10^2 \\
    c_1 &= 7.4959\cdot10^3    \\
    c_2 &= 1.2454\cdot10^6
    \end{align}
\end{subequations}}
\bb{where $C_H(\mu)$ and $C_Q(\mu)$ are $4^{th}$ order polynomial functions on the viscosity $\mu$ with coefficients defined in \cite{mheForESP}.}
\end{itemize}

\begin{table}[htbp]
\centering
\caption{ESP Model Variables}
\label{tab:ESP:ModelVariables}
\bb{\begin{tabular}{ll}
\hline
\multicolumn{2}{|c|}{Control inputs}                                             \\ \hline
\multicolumn{1}{|c|}{$f$}     & \multicolumn{1}{l|}{ESP frequency}                          \\ \hline
\multicolumn{1}{|c|}{$z$}     & \multicolumn{1}{l|}{Choke valve opening}                    \\ \hline
                            &                                                             \\ \hline
\multicolumn{2}{|c|}{ESP data}                                                   \\ \hline
\multicolumn{1}{|c|}{$p_m$}    & \multicolumn{1}{l|}{Production manifold pressure}           \\ \hline
\multicolumn{1}{|c|}{$p_{wh}$}   & \multicolumn{1}{l|}{Wellhead pressure}                      \\ \hline
\multicolumn{1}{|c|}{$p_{bh}$}   & \multicolumn{1}{l|}{Bottomhole pressure}                    \\ \hline
\multicolumn{1}{|c|}{$p_{p,in}$}  & \multicolumn{1}{l|}{ESP intakepressure}                     \\ \hline
\multicolumn{1}{|c|}{$p_{p,dis}$} & \multicolumn{1}{l|}{ESP discharge pressure}                 \\ \hline
\multicolumn{1}{|c|}{$p_r$}    & \multicolumn{1}{l|}{Reservoir pressure}                     \\ \hline
                            &                                                             \\ \hline
\multicolumn{2}{|c|}{Parameters from fluid analysis and well tests}              \\ \hline
\multicolumn{1}{|c|}{$q$}     & \multicolumn{1}{l|}{Average liquid flow rate}                      \\ \hline
\multicolumn{1}{|c|}{$q_r$}    & \multicolumn{1}{l|}{Flow rate from reservoir into the well} \\ \hline
\multicolumn{1}{|c|}{$q_c$}    & \multicolumn{1}{l|}{Flow rate through production choke}     \\ \hline
                            &          
\end{tabular}}
\end{table}

\bb{The state and algebraic variables involved in the ESP model appear in Table \ref{tab:ESP:ModelVariables}.
   The parameters used in this model are based on the parameters from \cite{mheForESP}. 
Table \ref{tab:ESP:modelParameters} presents the parameters which consist of fixed values such as well dimensions and ESP parameters, and parameters found from analysis of fluid such as bulk modulus $\beta_i$ and density $\rho$ \cite{mheForESP}. Parameters such as the well productivity index $PI$, viscosity $\mu$, and manifold pressure $p_m$ are assumed constant.}

\begin{table}[htbp]
\centering
\caption{ESP Model Parameters}
\label{tab:ESP:modelParameters}
\bb{\begin{tabular}{llll}
\hline
\multicolumn{4}{|c|}{Well dimensions and other known constants}             \\ \hline
\multicolumn{1}{|c|}{$g$}        & \multicolumn{1}{l|}{Gravitational acceleration constant}  & \multicolumn{1}{l|}{$9.81$}                        & \multicolumn{1}{l|}{$m/s^2$}    \\ \hline
\multicolumn{1}{|c|}{$C_c$}     & \multicolumn{1}{l|}{Choke valve constant}                 & \multicolumn{1}{l|}{$2\cdot10^{-5}$}    & \multicolumn{1}{l|}{*}                         \\ \hline
\multicolumn{1}{|c|}{$A_1$}     & \multicolumn{1}{l|}{Cross-section area of pipe below ESP} & \multicolumn{1}{l|}{0.008107}                    & \multicolumn{1}{l|}{$m^2$}      \\ \hline
\multicolumn{1}{|c|}{$A_2$}     & \multicolumn{1}{l|}{Cross-section area of pipe above ESP} & \multicolumn{1}{l|}{0.008107}                    & \multicolumn{1}{l|}{$m^2$}      \\ \hline
\multicolumn{1}{|l|}{$D_1$}     & \multicolumn{1}{l|}{Pipe diameter below ESP}              & \multicolumn{1}{l|}{0.1016}                      & \multicolumn{1}{l|}{$m$}                         \\ \hline
\multicolumn{1}{|l|}{$D_2$}     & \multicolumn{1}{l|}{Pipe diameter above ESP}              & \multicolumn{1}{l|}{0.1016}                      & \multicolumn{1}{l|}{$m$}                         \\ \hline
\multicolumn{1}{|l|}{$h_1$}     & \multicolumn{1}{l|}{Height from reservoir to ESP}         & \multicolumn{1}{l|}{200}                         & \multicolumn{1}{l|}{$m$}                         \\ \hline
\multicolumn{1}{|l|}{$h_w$}     & \multicolumn{1}{l|}{Total vertical distance in well}      & \multicolumn{1}{l|}{1000}                        & \multicolumn{1}{l|}{$m$}                         \\ \hline
\multicolumn{1}{|l|}{$L_1$}     & \multicolumn{1}{l|}{Length from reservoir to ESP}         & \multicolumn{1}{l|}{500}                         & \multicolumn{1}{l|}{$m$}                         \\ \hline
\multicolumn{1}{|l|}{$L_2$}     & \multicolumn{1}{l|}{Length from ESP to choke}             & \multicolumn{1}{l|}{1200}                        & \multicolumn{1}{l|}{$m$}                         \\ \hline
\multicolumn{1}{|l|}{$V_1$}     & \multicolumn{1}{l|}{Pipe volume below ESP}                & \multicolumn{1}{l|}{4.054}                       & \multicolumn{1}{l|}{$m^3$}      \\ \hline
\multicolumn{1}{|l|}{$V_2$}     & \multicolumn{1}{l|}{Pipe volume above ESP}                & \multicolumn{1}{l|}{9.729}                       & \multicolumn{1}{l|}{$m^3$}      \\ \hline
                               &                                                    
                               &                                                  &                                                \\ \hline
\multicolumn{4}{|c|}{ESP data}                                                                                                                                                        \\ \hline
\multicolumn{1}{|l|}{$f_0$}     & \multicolumn{1}{l|}{ESP characteristics reference freq.}  & \multicolumn{1}{l|}{60}                          & \multicolumn{1}{l|}{Hz}                        \\ \hline
\multicolumn{1}{|l|}{$I_{np}$}    & \multicolumn{1}{l|}{ESP motor nameplate current}          & \multicolumn{1}{l|}{65}                          & \multicolumn{1}{l|}{A}                         \\ \hline
\multicolumn{1}{|l|}{$P_{np}$}    & \multicolumn{1}{l|}{ESP motor nameplate power}            & \multicolumn{1}{l|}{$1.625\cdot10^5$} & \multicolumn{1}{l|}{W}                         \\ \hline
                               &                                                           &                                                  &                                                \\ \hline
\multicolumn{4}{|c|}{Parameters from fluid analysis and well tests}                                                                                                                   \\ \hline
\multicolumn{1}{|l|}{$\beta_1$}  & \multicolumn{1}{l|}{Bulk modulus below ESP}               & \multicolumn{1}{l|}{$1.5\cdot10^9$}   & \multicolumn{1}{l|}{Pa}                        \\ \hline
\multicolumn{1}{|l|}{$\beta_2$}  & \multicolumn{1}{l|}{Bulk modulus below ESP}               & \multicolumn{1}{l|}{$1.5\cdot10^9$}   & \multicolumn{1}{l|}{Pa}                        \\ \hline
\multicolumn{1}{|l|}{$M$}        & \multicolumn{1}{l|}{Fluid inertia parameter}              & \multicolumn{1}{l|}{$1.992\cdot10^8$} & \multicolumn{1}{l|}{$kg/m^4$}   \\ \hline
\multicolumn{1}{|l|}{$\rho$} & \multicolumn{1}{l|}{Density of produced fluid}            & \multicolumn{1}{l|}{950}                         & \multicolumn{1}{l|}{$kg/m^3$}   \\ \hline
\multicolumn{1}{|l|}{$P_r$}     & \multicolumn{1}{l|}{Reservoir pressure}                   & \multicolumn{1}{l|}{$1.26\cdot10^7$}  & \multicolumn{1}{l|}{Pa}                        \\ \hline
                               &                                                           &                                                  &                                                \\ \hline
\multicolumn{4}{|c|}{Parameters assumed to be constant}                                                                                                                               \\ \hline
\multicolumn{1}{|l|}{PI}       & \multicolumn{1}{l|}{Well productivity index}              & \multicolumn{1}{l|}{$2.32\cdot10^{-9}$} & \multicolumn{1}{l|}{$m^3/s/$Pa} \\ \hline
\multicolumn{1}{|l|}{$\mu$}       & \multicolumn{1}{l|}{Viscosity of produced fluid}          & \multicolumn{1}{l|}{0.025}                       & \multicolumn{1}{l|}{$Pa \cdot s$}                      \\ \hline
\multicolumn{1}{|l|}{$P_m$}       & \multicolumn{1}{l|}{Manifold pressure}                    & \multicolumn{1}{l|}{20}                          & \multicolumn{1}{l|}{Pa}                        \\ \hline
\end{tabular}}
\end{table}





\end{document}